\documentclass[10pt,twocolumn,letterpaper]{article}

\usepackage{3dv}
\usepackage{times}
\usepackage{epsfig}
\usepackage{graphicx}
\usepackage{amsmath}
\usepackage{amssymb}
\usepackage{array}
\usepackage{pifont}
\usepackage{capt-of,etoolbox}
\usepackage{multirow}
\usepackage{cite}
\usepackage{hhline}
\usepackage{authblk}
\usepackage{bbding}
\usepackage{algorithm}
\usepackage{algorithmic}
\usepackage{gensymb}
\usepackage[utf8x]{inputenc} 

\usepackage[pagebackref=true,breaklinks=true,colorlinks,bookmarks=false]{hyperref}

\threedvfinalcopy 

\def\threedvPaperID{150} 

\ifthreedvfinal\fi
\begin{document}

\title{Synergy between 3DMM and 3D Landmarks for Accurate 3D Facial Geometry}

\author{Cho-Ying Wu \qquad Qiangeng Xu \qquad Ulrich Neumann\\
University of Southern California\\
}

\maketitle
\thispagestyle{empty}

\begin{abstract}

This work studies learning from a synergy process of 3D Morphable Models (3DMM) and 3D facial landmarks to predict complete 3D facial geometry, including 3D alignment, face orientation, and 3D face modeling. Our synergy process leverages a representation cycle for 3DMM parameters and 3D landmarks. 3D landmarks can be extracted and refined from face meshes built by 3DMM parameters. We next reverse the representation direction and show that predicting 3DMM parameters from sparse 3D landmarks improves the information flow. Together we create a synergy process that utilizes the relation between 3D landmarks and 3DMM parameters, and they collaboratively contribute to better performance. We extensively validate our contribution on full tasks of facial geometry prediction and show our superior and robust performance on these tasks for various scenarios. Particularly, we adopt only simple and widely-used network operations to attain fast and accurate facial geometry prediction. Codes and data: \url{ https://choyingw.github.io/works/SynergyNet/}.
\vspace{-14pt}
\end{abstract}

\section{Introduction}

\label{intro}
Facial geometry prediction including 3D facial alignment, face orientation estimation, and 3D face modeling are fundamental tasks \cite{zhang2016joint, zollhofer2018state, chang2019deep, tuan2018extreme, kim2018deep, wu2019mvf,deng2019accurate,tuan2017regressing,lv2017deep,wu2022cross} and have applications on face recognition \cite{shi2006effective, wu2018occluded, wu2016occlusion, wu2019efficient, deng2021masked, juhong2017face}, tracking \cite{liu2017robust, deng2019menpo, lin2018prior,chrysos2018comprehensive}, and compression \cite{wang2020one}. Recent works \cite{zhu2016face, zhu2019face, guo2020towards, tu20203d, feng2018joint} predict facial geometry by estimating 3D Morphable Models (3DMM) parameters that include shape and expression variations. Yet, face orientation for these previous works is only a by-product without evaluation and discussion on relations with 3D landmarks and 3D face models. In contrast, we fully evaluate facial geometry, including 3D facial alignment, face orientation estimation, and 3D face modeling.

3D face meshes can be built from 3DMM parameters, and 3D landmarks can be extracted from vertices by querying associated indices. 3D landmarks are widely used to guide 3D facial geometry learning. Previous works \cite{zhu2016face, zhu2019face, tu20203d, guo2020towards, feng2018joint} only directly extract coarse landmarks from fitted 3D faces and compute supervised alignment losses with groundtruth landmarks. These works utilize a representation direction, from 1D parameters to 3D landmarks. However, though 3D landmarks are very sparse (a 68-point definition is commonly used), they compactly and efficiently describe facial outlines in 3D space. We think 3D landmarks can be further exploited to predict underlying 3D faces as supportive information. Hence, in addition to only going from 1D parameters to 3D landmarks, we propose a further step to reversely regress 3DMM parameters from 3D landmarks and establish a representation cycle. The advantage is that \textit{predicting 3D face using 3DMM from 2D images is naturally an ill-posed problem, but prediction from 3D landmarks can alleviate the intrinsic ill-posedness}. To our knowledge, we are the first to study this reverse representation direction, from 3D landmarks to 3DMM parameters. Together we build a representation cycle as a \textit{synergy process} that adopts collaborative relation between 3DMM parameters and 3D landmarks to improve the information flow and attain better performance.


\begin{figure}[bt!]
\begin{center}
\includegraphics[width=1.0\linewidth]{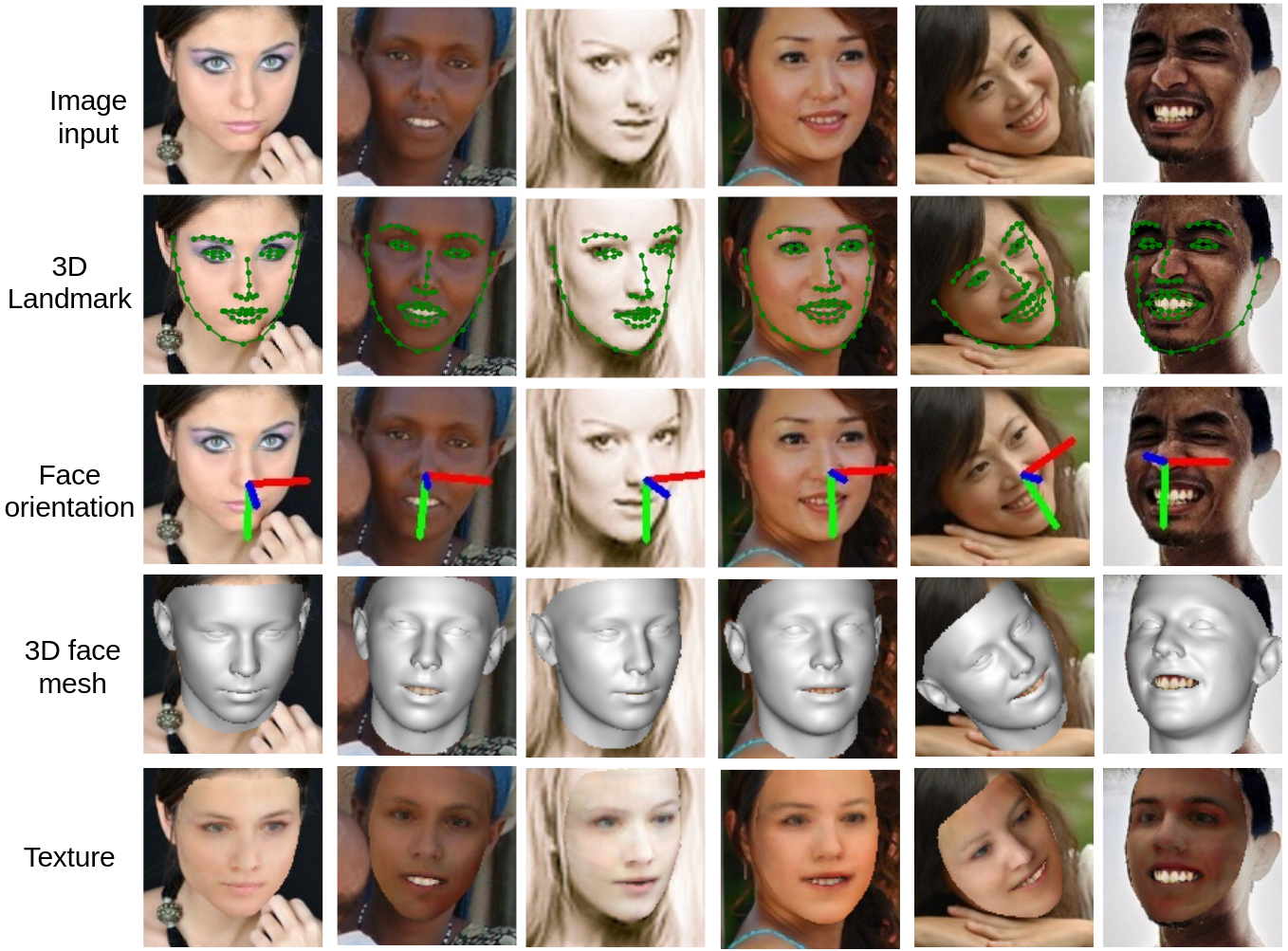}
\end{center}
\vspace{-13pt}
   \caption{\textbf{Results from our SynergyNet with monocular image inputs.} Note that 3D landmarks can predict hidden face outlines in 3D rather than follow visible outlines on images.}
   \vspace{-18pt}
\label{teaser}
\end{figure}

\begin{figure}[bt!]
\begin{center}
\includegraphics[width=1.0\linewidth]{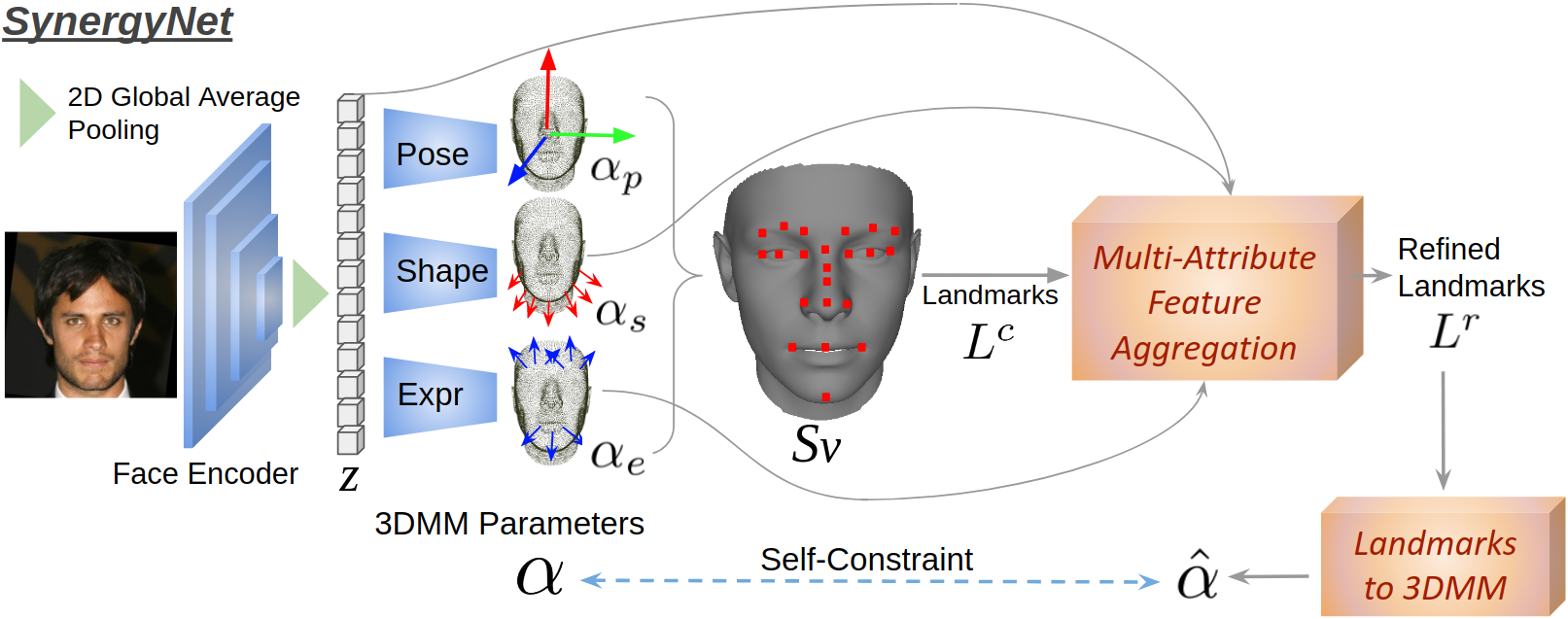}
\end{center}
\vspace{-12pt}
   \caption{\textbf{Framework of our SynergyNet.} Backbone network learns to regress 3DMM parameters ($\alpha_p$,$\alpha_s$, and $\alpha_e$) and reconstruct 3D face meshes from monocular face images. Multi-Attribute feature aggregation gathers underlying 3DMM semantics and the latent image code to refine landmarks further. The landmark-to-3DMM module regresses 3DMM from refined landmarks $L^r$ to reveal the embedded facial geometry in 3D landmarks. A self-constraining consistency is applied to 3DMM parameters regressed from different sources. This synergy process includes a forward representation direction, from \textit{3DMM parameters to refined 3D landmarks}, and a reverse direction, from \textit{3D landmarks to regress 3DMM parameters}, to attain better performance. The red and blue arrows after shape and expression (expr) decoders show the main areas of deformation that each 3DMM semantics controls.}
   \vspace{-8pt}
\label{pipeline}
\end{figure}

We propose \textbf{SynergyNet}, a synergy process network that includes two stages. The first stage contains a backbone network to regress 3DMM parameters from images and construct 3D face meshes. After landmark extraction by querying associated indices, we propose a landmark refinement module that aggregates 3DMM semantics and incorporates them into point-based features to produce refined 3D landmarks. We closely validate how each information source contributes to 3D landmark refinement. From the representation perspective, the first stage goes from 1D parameters to 3D landmarks. Next, the second stage contains a landmark-to-3DMM module that predicts 3DMM parameters from 3D landmarks, which is a reverse representation direction compared with the first stage. We leverage this step to regress embedded facial geometry lying in sparse landmarks. The overall framework is in Fig. \ref{pipeline}.

We will first review 3DMM as basics in Sec.\ref{itemized3DMM} used in the previous work \cite{zhu2016face, zhu2019face,tu20203d,guo2020towards}. 3DMM regression contains pose, shape, and expression parameter estimation from a monocular face image through a backbone network. 3D faces are constructed as foundation models by 3DMM, and 3D landmarks are extracted from face meshes. Next, in Sec.\ref{landmarkRef}, we introduce the proposed multi-attribute feature aggregation (MAFA), including landmark features, image features, and shape and expression of 3DMM semantics. MAFA is then used to produce finer landmark structures. The advantage is that refinement based on only coarse landmarks is hard because the information is unitary. Joining different attributes can refine and correct raw structures.

In Sec.\ref{pgs}, we introduce the reverse direction to regress 3DMM parameters from 3D landmarks, based on the assumption that 3D landmarks contain rough facial geometry. The advantage is that regressing 3DMM parameters from 3D landmarks can avoid inherent ambiguity in conventional 3DMM-based methods that predict facial geometry only from images. A self-constraining loss for both 3DMM parameters regressed from images and from 3D landmarks is used: since two 3DMM parameters describe the same identity, they should be numerically consistent. 



Especially, our SynergyNet contains only simple and widely-used network operations in the whole synergy process. We quantitatively analyze performance gains introduced by each adopted information and each regression target with extensive experiments. We evaluate our SynergyNet on all tasks of facial alignment, face orientation estimation, and 3D face modeling using the standard datasets for each task. Our SynergyNet attains superior performance than other related work. Fig.\ref{teaser} demonstrates the ability of our SynergyNet.

In summary, we present the following contributions:

1. We propose SynergyNet to study a synergy process that leverages the collaborative relation between 3DMM parameters and 3D landmarks to learn better 3D facial geometry. This is the first study to include reverse representation direction, i.e., from 3D landmarks to 3DMM parameters.

2. We propose multi-attribute feature aggregation for landmark refinement using multiple information sources and closely analyze performance gain for each information.  

3. We conduct extensive and detailed benchmarking on 3D facial geometry, including facial alignment, face orientation estimation, and 3D face modeling, to validate our superior performance on these tasks.

\section{Related Work}
\subsection{3D Facial Alignment via 3D Face Modeling}
\vspace{-5pt}
3D information is useful in wide vision applications~\cite{wu2022toward,wu2023inspacetype,wu2023meta,wu2021scene,zhong2019deep}.
3D facial alignment aims at predicting 3D landmarks on images. In contrast, 2D approaches \cite{kazemi2014one, dong2018style, dong2019teacher,feng2018wing, wu2018look} usually regress direct landmark coordinates or heatmaps based on \textit{visible} facial parts. If input faces are self-occluded due to large face poses, their methods either only estimate landmarks along visible face outlines rather than hidden outlines or produce much larger errors at invisible parts that make the results unreliable.  

3D approaches \cite{zhu2016face, zhu2019face, feng2018joint, bulat2017far, guo2020towards, Jackson_2017_ICCV, shang2020self} predict aligned 3D faces with images. This way, occluded landmarks can be registered. 3DDFA \cite{zhu2016face, zhu2019face} adopts Basel Face Model (BFM) and use 3DMM fitting to reconstruct face meshes from monocular images. PRNet \cite{feng2018joint} predicts 2D UV-position maps that encode 3D points and uses BFM mesh connectivity to build face models. Compared with 3DDFA, PRNet might have higher mesh deformation ability since its 3D points are not from 3DMM parameterization. However, it is harder to obtain a smooth and reliable mesh for PRNet. 2DASL \cite{tu20203d} based on 3DMM further adopts a differentiable renderer and a discriminator to produce high-quality 3D face models. 3DDFA-V2 \cite{guo2020towards} based on 3DDFA further introduces a meta-joint optimization strategy and a short video synthesis to attain the current best result. 3D faces and 3D landmarks extracted by vertex indexing are outputs of these methods. However, their landmarks are raw without refinement. Our landmarks are refined with multi-attribute feature aggregation, and we further adopt 3DMM from 3D landmarks as another information source.

Another line of work adopts self-supervision from images and targets at more realistic 3D face synthesis \cite{tewari2018self,tewari2019fml,feng2021learning,sanyal2019learning, wen2021self}. Their self-supervised factors usually rely on visible facial areas to collect visual cues for prediction and may not be robust to large-pose cases.

\subsection{Face Orientation Estimation}
\vspace{-5pt}
Face orientation estimation has applications on human-robot interaction \cite{lemaignan2016real,wang2018human,palinko2016robot}. Euler angles (yaw, pitch, roll) are used to represent the orientation. Deep Head Pose \cite{mukherjee2015deep} uses networks to predict 2D landmarks and face orientation at the same time. HopeNet \cite{ruiz2018fine} uses bin-based angle regression and QuatNet \cite{hsu2018quatnet} uses a multi-regression loss for head pose. FSA-Net \cite{yang2019fsa} constructs a fine-grained structure mapping for features aggregation. TriNet \cite{cao2021vector} uses a vector-based representation for pose estimation. These works focus on face orientation as a standalone task. On the other hand, although 3DMM-based 3D alignment approaches estimate rotation matrices, previous works only focus on evaluation and discussion on landmarks and 3D faces \cite{zhu2016face, zhu2019face, tu20203d, guo2020towards}. To gain an insight into full facial geometry, we benchmark both standalone orientation estimation methods and 3DMM-based approaches.

\section{Method}
Our method, illustrated in Fig. \ref{pipeline}, aims at precise and accurate 3D facial alignment, face orientation estimation, and 3D face modeling by utilizing a synergy process of 3D landmarks and 3DMM parameters to guide 3D facial geometry learning better. The pipeline contains two stages. The first stage includes a preliminary 3DMM regression from images and a multi-attribute feature aggregation (MAFA) for landmark refinement. The second stage contains a landmark-to-3DMM regressor to reveal the embedded facial geometry in sparse landmarks.

\subsection{3D Morphable Models (3DMM)}
\label{itemized3DMM}
3DMM reconstructs face meshes using principal component analysis (PCA). Given a mean face $M \in \mathbb{R}^{3N_v}$ with $N_v$ 3D vertices, 3DMM deforms $M$ into a target face mesh by predicting the shape and expression variations. $U_{s}\in \mathbb{R}^{3N_v\times 40}$ is the basis for shape variation manifold that represents different identities, $U_{e}\in \mathbb{R}^{3N_v\times 10}$ is the basis for expression variation manifold, and $\alpha_{s}\in \mathbb{R}^{40}$ and $\alpha_{e}\in \mathbb{R}^{10}$ are the associated basis coefficients. The 3D face reconstruction can be formulated in Eq.\ref{3DMM_basics}.
\begin{equation}
     S_f = Mat(M + U_{s}\alpha_{s} + U_{e}\alpha_{e}),
\label{3DMM_basics}
\vspace{-3pt}
\end{equation}
where $S_{f}\in \mathbb{R}^{3\times N_v}$ represents a reconstructed frontal face model after the vector-to-matrix operation ($Mat$). To align $S_{f}$ with input view, a 3x3 rotation matrix $R\in SO(3)$, a translation vector $t\in \mathbb{R}^3$, and a scale $\tau$ are predicted to transform $S_f$ by Eq.\ref{affine_trans}.
\begin{equation}
	S_v = \tau RS_f+t,
\label{affine_trans}
\vspace{-3pt}
\end{equation}
where $S_v\in \mathbb{R}^{3\times N_v}$ aligns with input view. $\tau R$ and $t$ are included as 3DMM parameters in most works \cite{wu2019mvf, zhu2016face, guo2020towards}, and thus we use $\alpha_{p}\in \mathbb{R}^{12}$ instead. We follow the current best work 3DDFA-V2 \cite{guo2020towards} to predict 62-dim 3DMM parameters $\alpha$ for pose, shape, and expression.

We follow 3DDFA-V2 to adopt MobileNet-V2 as the backbone network to encode input images and use fully-connected (FC) layers as decoders for predicting 3DMM parameters from the bottleneck image feature $z$. We separate the decoder into several heads by 3DMM semantics, which jointly predict the whole 62-dim parameters. The advantage of separate heads is that disentangling pose, shape, and expression controls secures better information flow. The illustration in Fig. \ref{pipeline} shows the encoder-decoder structure. The decoding is formulated as $\alpha_{m}=\text{Dec}_{m}(z)$, $m\in\{p, s, e\}$, showing pose, shape and expression. With groundtruth notation $^*$ hereafter, the supervised 3DMM regression loss is shown as follows.
\vspace{-7pt}
\begin{equation}
     \mathbb{L}_{\text{3DMM}} = \sum_{m} \|\alpha_{m}-\alpha^*_{m}\|^2.
\label{loss_S1_3dmm}
\vspace{-7pt}
\end{equation}

\subsection{From 3DMM to Refined 3D Landmarks}
\label{landmarkRef}
After regressing 3DMM parameters ($\alpha_p$,$\alpha_s$,$\alpha_e$), 3D face mesh for the input face can be constructed by Eq.\ref{3DMM_basics} and be aligned with input face by Eq.\ref{affine_trans}. We adopt popular BFM \cite{paysan20093d}, which includes about 53K vertices, as the mean face $M$ in Eq.\ref{3DMM_basics}. Then, 3D landmarks $L^c \in \mathbb{R}^{3 \times N_l}$ are extracted by landmark indices. $N_l=68$ is used in 300W-LP\cite{zhu2016face} as our training dataset. 


Previous studies \cite{guo2020towards, zhu2019face, zhu2016face} directly use extracted landmarks $L^c$ to compute the alignment loss for learning 3D facial geometry. However, these extracted landmarks are raw without refinement. Instead, we adopt a refinement module that aggregates multi-attribute features to produce finer landmark structures. Landmarks can be seen as a sequence of 3D points. Weight-sharing multi-layer perceptrons (MLPs) are commonly used for extracting features from structured points. PointNet-based frameworks \cite{qi2017pointnet, qi2017pointnet++,wang2019pseudo,ma2020rethinking,xu2020grid,wu2020geometry} use an MLP-encoder to extract high-dimensional embeddings. At the bottleneck, global point max-pooling is applied to obtain global point features. Then an MLP-decoder is used to regress per-point attributes. An MLP-based refinement module takes sparse landmarks $L^c$ as inputs and uses the MLP-encoder and MLP-decoder to produce finer landmarks.

\begin{figure}[bt!]
    \centering
    \includegraphics[width=0.93\linewidth]{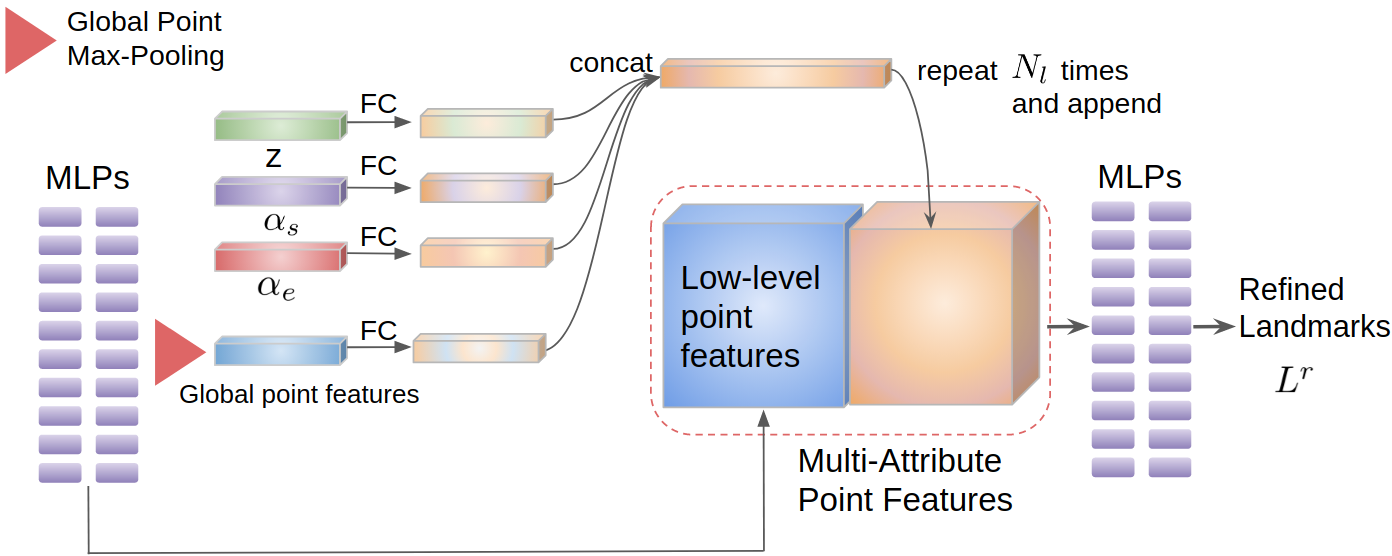}
    \vspace{-3pt}
    \caption{\textbf{Structure of multi-attribute landmark refinement.} The input is $L^c$ from the foundation face model. The left MLPs extract global point features and fuse the global features with other attributes, including images features, shape, and expression parameters. The concatenation is appended to the low-level features to create multi-attribute point features, which are used to regress the refined landmarks.}
    \vspace{-15pt}
    \label{mmfv}
\end{figure}

Instead of using $L^c$ alone for the refinement, our refinement module adopts multi-attribute feature aggregation (MAFA), including input images and 3DMM semantics that provide information from different domains. For example, shape contains information of thinner/thicker faces, and expression contains information for eyebrow or mouth movements. Therefore, these pieces of information can help regress finer landmark structures. Specifically, our MAFA fuses information of the image, using its bottleneck features $z$ after global average pooling and shape and expression 3DMM parameters. These features and parameters are global information without spatial dimensions. We first use FC layers for domain adaption. Later we concatenate them into a multi-attribute feature vector and then repeat this vector $N_l$ times to make multi-attribute features compatible with per-point features. We last append the repeated features to the low-level point features and feed them to an MLP-decoder to produce refined 3D landmarks.
The overall design is shown in Fig. \ref{mmfv}. Skip connection is used from the coarse to refined landmarks to facilitate training. 

We use groundtruth landmarks to guide the training. The alignment loss function is formulated as follows.
\begin{equation}
     \mathbb{L}_{lmk} = \sum_{n} \mathbb{L}_{smL1}(L^r_n-L_n^*), n\in[1,N_l],
\label{loss_S1_lmk}
\vspace{-5pt}
\end{equation}
where $N_l$ is number of landmarks, $^*$ denotes groundtruth, and $\mathbb{L}_{smL1}$ is smooth L1 loss. So far, the operations of constructing 3D face meshes and landmark extraction and refinement transform 3DMM parameters to refined 3D landmarks.

\subsection{From Refined Landmarks to 3DMM} 
\label{pgs}
We next describe the reverse direction of representation that goes from refined landmarks to 3DMM parameters. 

Previous works only consider 3DMM parameter regression from images \cite{zhu2016face, zhu2019face,guo2020towards,tu20203d, deng2019accurate,wu2019mvf,Jackson_2017_ICCV}. However, facial landmarks are sparse keypoints lying at eyes, nose, mouth, and face outlines, which are principal areas that $\alpha_s$ and $\alpha_e$ control. We assume that approximate facial geometry is embedded in sparse landmarks. Thus, we further build a landmark-to-3DMM module to regress 3DMM parameters from the refined landmarks $L^r$ using the holistic landmark features. To our knowledge, we are the first to study this reverse representation direction, from landmarks to 3DMM parameters.

The landmark-to-3DMM module also contains an MLP-encoder to extract high dimensional point features and use a global point max-pooling to obtain holistic landmark features. Later separate FC layers transform the holistic landmark features to 3DMM parameters to get $\hat{\alpha}$, including pose, shape, and expression. We refer $\hat{\alpha}$ to \textbf{landmark geometry}, since this 3DMM geometry is regressed from landmarks. We adopt a supervised loss with groundtruth $\alpha^*$ for $\hat{\alpha}$ as follows. 
\begin{equation}
     \mathbb{L}_{\text{3DMM}_{lmk}} = \sum_{m} \|\hat{\alpha}_{m}-\alpha^*_m\|^2,
\label{loss_S2_3dmm}
\vspace{-7pt}
\end{equation}
where $m$ contains pose, shape, and expression.

Furthermore, since $\hat{\alpha}$ regressed from the landmarks and $\alpha$ regressed from the face image describe the same identity, they should be numerically similar. We further add a novel self-supervision control as follows.
\begin{equation}
\small
     \mathbb{L}_{g} = \sum_{m} \|\alpha_{m}-\hat{\alpha}_{m}\|^2,
\label{loss_S2_corr}
\vspace{-7pt}
\end{equation}
where $m \in \{p, s, e\}$. $\mathbb{L}_g$ improves information flow that lets 3DMM regressed from images obtain support from landmark geometry.

The advantage of self-supervision control (Eq.\ref{loss_S2_corr}) is that since images and sparse landmarks are different data representations (2D grids and 3D points) using different network architectures and operations, more descriptive and richer features can be extracted and aggregated under this multi-representation strategy. Although conceptually sparse landmarks provide rough face outlines, our experiments show that this reverse representation direction further contributes to the performance gain and attains superior results than related work. 

Overall, the total loss combination is shown as follows
\begin{equation}
\small
     \mathbb{L}_{total} = \lambda_1\mathbb{L}_{\text{3DMM}}+\lambda_2\mathbb{L}_{lmk}+\lambda_3\mathbb{L}_{\text{3DMM}_{lmk}}+\lambda_4\mathbb{L}_{g},
\label{loss_total}
\vspace{-3pt}
\end{equation}
where $\lambda$ terms are loss weights. 



\subsection{Representation Cycle}
\label{cycle}
\begin{figure}[tb!]
    \centering
    \includegraphics[width=0.85\linewidth]{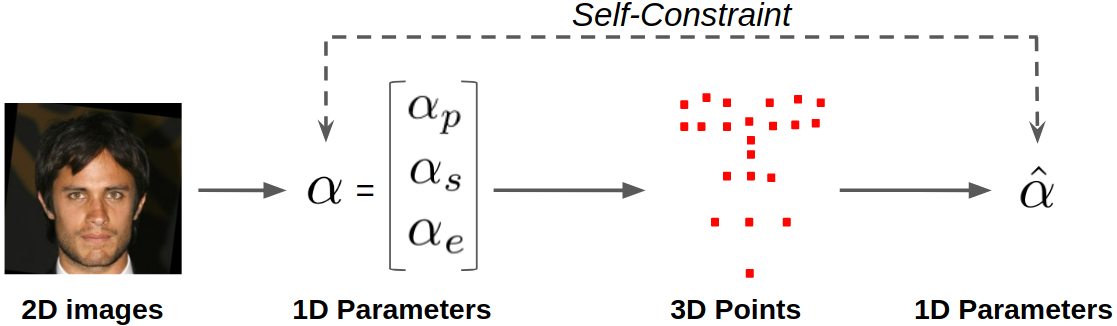}
    \caption{\textbf{Illustration of representation cycle.} }
    \label{representation_cycle}
    \vspace{-17pt}
\end{figure}

Our overall framework creates a cycle of representations. First, the image encoder and separate decoders regress 1D parameters from a face image input. Then, we construct 3D meshes from parameters and refine extracted 3D landmarks—this is the forward direction that switches representations from 1D parameters to 3D points. Next, the reverse representation direction adopts a landmark-to-3DMM module to switch representations from 3D points back to 1D parameters. Therefore it forms a representation cycle (Fig. \ref{representation_cycle}), and we minimize the consistency loss to facilitate the training. The forward and reverse representation direction between 3DMM parameters and refined 3D landmarks form a synergy process that collaboratively improves the learning of facial geometry. Landmarks are extracted and refined, and the refined landmarks and landmark geometry further supports better 3DMM parameter predictions using the self-supervised consistency loss (Eq.\ref{loss_S2_corr}).

Compared with a simple baseline using only the forward representation, i.e., going from image to 3DMM and directly extracting 3D points from built meshes to compute alignment loss, our proposed landmark refinement (MAFA) and the reverse representation (landmark-to-3DMM module) only bring about 5\% more time in average for a single feed-forward pass. This is because landmarks are sparse and compact, and weight-sharing MLPs are lightweight. 

We choose simple and widely-used network operations to show that without special operations, landmarks and 3DMM parameters can still guide the 3D facial geometry learning better. Through the following studies and experiments, we closely validate each module we introduce to the plain 3DMM regression from images, including MAFA for landmark refinement and landmark-to-3DMM module. Network details are described in the supplementary.

\section{Experiments}
\label{exp}
Evaluation is conducted on the three focused tasks: facial alignment, face orientation estimation, and 3D face models. One of our main contributions is close studies that exhibit a detailed performance gain breakdown for each module we propose and each attribute we include in MAFA.

\textbf{Procedure.}
We train on 300W-LP\cite{zhu2016face}, which is a standard and widely-used training set on 3D face tasks. It collects in-the-wild face images and conducts face profiling \cite{zhu2016face} for producing 3DMM parameters with BFM \cite{paysan20093d} and FaceWarehouse texture \cite{cao2013facewarehouse}. The dataset also performs out-of-plane face rotation for augmentation, attaining more than 60K face images and fitted 3D models.

During training, we use a learning rate of 0.08, batch size of 1024, and momentum of 0.9 of the SGD \cite{zinkevich2010parallelized} optimizer. We train our network with 80 epochs, and the learning rate decays to $\frac{1}{10}$ and $\frac{1}{100}$ of the initial after 48 and 64 epochs. We use random color jittering and random horizontal flip. We further adopt face-swapping augmentation that exchanges textures with others, overlays them on the original mesh, and last renders the 3D model onto the original image. This is to increase the appearance variety that creates novel texture-geometry pairs. We train on 4x GTX 1080Ti GPUs, and the training takes about 8 hours.



At test time, the refined landmarks $L^r$, fitted 3D face $S_v$ with its orientation from $\alpha_p$ are the outputs for evaluation. The processing of landmark geometry is saved at test time since its information is auxiliary, and we leave the discussion and evaluation of landmark geometry in the supplementary. Our inference attains 2600fps inference speed on average for the 3D landmark prediction and about 2300fps for the dense 3D face prediction on a single GPU with the MobileNet backbone. The speed calculation includes inference time batching and communication overhead of data loading to GPU-memory. This satisfies the real-world applications for fast inference. 

In addition to facial geometry, we also study texture synthesis in the supplementary for more realistic 3D faces.

\begin{table*}[tb!]
\begin{center}
  \caption{\textbf{Ablative for facial alignment.} The first table is for AFLW2000-3D using original groundtruth annotation. The second table is for the reannotated version. '-' means the module is not used, and the corresponding loss terms are not introduced. The first row setting without all the introduced modules contains only a simple baseline of a backbone network to regress 3DMM parameters from only images.}
  \label{FAL_ab}
  \footnotesize
  \begin{tabular}[c]
  {|p{2.0cm}<{\centering\arraybackslash}|
  p{3.4cm}<{\centering\arraybackslash}|
  p{2.2cm}<{\centering\arraybackslash}|
  p{1.0cm}<{\centering\arraybackslash}|
  p{1.0cm}<{\centering\arraybackslash}|
  p{1.0cm}<{\centering\arraybackslash}|
  p{1.0cm}<{\centering\arraybackslash}|
  p{0.7cm}<{\centering\arraybackslash}|}
  \hline
  
    AFLW2000-3D Original  & Multi-Attribute Feature Aggregation for Refinement & Landmark-to-3DMM  & 0 to 30 & 30 to 60 & 60 to 90 & All \\
    \hline
     &  - & -  & 2.99 & 3.80 & 4.86 & 3.88\\ 
     &  \Checkmark & -  & 2.68 & 3.32 & 4.35 & 3.49\\ 
     &  - & \Checkmark  & 2.69 & 3.57 & 4.69 & 3.65\\ 
     &  \Checkmark & \Checkmark &  \textbf{2.66} & \textbf{3.30} & \textbf{4.27} & \textbf{3.41}\\
    \hline
    \hline
    AFLW2000-3D Reannotated  & Multi-Attribute Feature Aggregation for Refinement & Landmark-to-3DMM & 0 to 30 & 30 to 60 & 60 to 90 & All \\
    \hline
     &  - & -  & 2.34 & 2.99 & 4.27 & 3.20\\ 
     &  \Checkmark & -  & 2.24 & 2.67 & 3.76 & 2.89\\ 
     &  - & \Checkmark & 2.23 & 2.69 & 3.90 & 2.94\\
     &  \Checkmark & \Checkmark &  \textbf{2.16} & \textbf{2.61} & \textbf{3.66} & \textbf{2.81}\\ 
    \hline
  \end{tabular}
  \vspace{-25pt}
\end{center}
\end{table*}

\textbf{Test sets for facial alignment.}
Facial alignment is standardly evaluated on AFLW2000-3D \cite{zhu2016face}, which contains the first 2000 images of AFLW \cite{koestinger11a} with a \textit{68-point} landmark annotation. Two landmark sets of AFLW2000-3D are present, original and reannotated by LS3D-W \cite{bulat2017far}). The reannotated one carries better quality. We separately report performances on the two versions to fairly compare with related work. We also follow \cite{guo2020towards} to evaluate on the full AFLW set, which contains 21K images with a \textit{21-point} landmark annotation. The two datasets are used for showing evaluation on different numbers of facial landmarks. 

\textbf{Test sets for 3D face modeling.}
We evaluate 3D face modeling on AFLW2000-3D\cite{zhu2016face}, MICC Florence\cite{florence2011}, 300VW \cite{shen2015first}, and Artistic-Faces \cite{yaniv2019face} for both quantitative and qualitative analysis. AFLW2000-3D contains 2000 fitted 3D faces. Florence contains high-resolution real face scans of 53 individuals. 300VW collects talks or interviews from the web, and Artistic-Faces gathers artistic style faces. 

\textbf{Test sets for face orientation estimation.}
Most previous 3DMM-based works \cite{zhu2016face, feng2018joint, guo2020towards, guo2020towards, bulat2017far} focus on the facial alignment and 3D face modeling. To fully evaluate facial geometry, we introduce face orientation estimation for evaluation. AFLW2000-3D contains large-pose faces with orientation annotation that make it suitable for evaluation.

\subsection{Facial Alignment Evaluation}
\label{FAL_section}

\textbf{Metrics.} Normalized mean error (NME) in Eq.\ref{NME_all} for sparse facial landmarks with Euclidean distance is reported.
\begin{equation}
\vspace{-2pt}
     \text{NME} = \frac{1}{T}\sum^T_{t=1}\frac{\|u_t-v_t\|_2}{B},
\label{NME_all}
\vspace{-2pt}
\end{equation}
where $u_t$ and $v_t$ are landmark prediction/groundtruth that are both registered with face images, $T$ is the number of samples, and $B$ is bounding box size, square root of box areas, as the normalization term for each face.

\textbf{Ablation study.} We show three different ablation studies in this section to thoroughly examine the contribution of each part in SynergyNet. The aim is to validate our introduced multi-attribute feature fusion and analyze how each attribute contributes to the final performance.

\textbf{(1)} We first conduct ablation studies of SynergyNet on AFLW2000-3D. Compared with conventional frameworks that only use a backbone network to regress 3DMM parameters from images, the proposed multi-attribute feature aggregation for landmark refinement and the landmark-to-3DMM module in the synergy process are examined. Following \cite{zhu2019face, guo2020towards, feng2018joint}, we report NME under three yaw angle ranges. The results are shown in Table \ref{FAL_ab}.

The table shows that both landmark refinement and landmark geometry support stages contribute to the final performance for better facial landmark estimation. MAFA adopts the advantage of multi-attribute information fusion to refine landmarks. The landmark-to-3DMM reveals the geometric information embedded in the 3D landmarks and helps 3DMM prediction with the representation cycle. From both tables' third and fourth rows, directly predicting landmark geometry from raw landmarks without refinement only obtains limited performance gains. By contrast, based on finer landmarks, the reverse representation stage further improves the results. This validates our SynergyNet design that both MAFA and the landmark-to-3DMM are required to attain the best performance.

\textbf{(2)} We then study the performance contribution of \textit{each attribute at MAFA} and \textit{each 3DMM regression target} at the reverse representation direction stage. For the former, we experiment with different feature aggregation and fusion: only point feature, point+image feature, and all attributes in Fig. \ref{mmfv} (point, image, and 3DMM semantics). For the latter, we examine the performance gain of each regression target at the landmark-to-3DMM, including pose, shape, and expression from 3D landmarks. Results are shown in Table \ref{FAL_PIFV_ab}. Row 1- 3 show that the gain of using image and 3DMM semantics mainly comes from small or medium pose ranges because images and the derived 3DMM parameters capture more descriptive features on frontal faces. Large poses may cause \textit{self-occlusion} on images and make prediction unreliable. Row 4 to 6 exhibit effects of regressing pose only (Row 4), shape and expression (Row 5), and all (Row 6). The improvements mainly come from large-pose cases. The reason is that the reverse direction regresses parameters from 3D landmarks that provide features from the 3D space, which naturally avoids self-occlusion compared with 2D. Such a strategy benefits alignment for large poses.

\begin{table}[tb!]
\begin{center}
  \caption{\textbf{Study on different attributes used at landmark refinement and different regression targets at landmark-to-3DMM (L$\rightarrow$3D).} AFLW2000-3D Original is adopted for facial alignment evaluation. The first three rows exploit different attributes for feature aggregation. Row 3 uses all attributes shown in Fig.\ref{mmfv}. Row 4 to 6 further study performance gains of different regression targets at the reverse representation direction.} \vspace{-5pt}
  \label{FAL_PIFV_ab}
  \footnotesize
  \begin{tabular}[c]
  {|
  p{3.25cm}<{\arraybackslash}|
  p{0.77cm}<{\centering\arraybackslash}|
  p{0.92cm}<{\centering\arraybackslash}|
  p{0.91cm}<{\centering\arraybackslash}|
  p{0.53cm}<{\centering\arraybackslash}|}
  \hline
      Structures  & 0 to 30 & 30 to 60 & 60 to 90 & All \\
    \hline
       Point feature only & 2.73 & 3.51 & 4.51 & 3.58\\ 
       Point + image feature & 2.68 & 3.40 & 4.61 & 3.56\\ 
       MAFA & 2.67 & 3.34 & 4.51 & 3.51\\ 
       MAFA+ L$\rightarrow$3D(pose) & 2.68 & 3.31 & 4.55 & 3.51 \\
       MAFA+ L$\rightarrow$3D(shape,expr) & 2.67 & 3.32 & 4.47 & 3.48 \\
       MAFA+ L$\rightarrow$3D(all) &  \textbf{2.66} & \textbf{3.30} & \textbf{4.27} & \textbf{3.41}\\ 
    \hline
  \end{tabular}
  \vspace{-21pt}
\end{center}
\end{table}

\begin{table}[tb!]
\begin{center}
  \caption{\textbf{Landmark-to-3DMM network structure study.} Facial alignment on AFLW2000-3D Original is evaluated. Refer to Section \ref{FAL_section} for the two comparison settings. The last row is the adopted setting introduced in Sec.\ref{pgs}.}
  \label{PGS_study}
  \footnotesize
  \begin{tabular}[c]
  {|
  p{2.3cm}<{\centering\arraybackslash}|
  p{0.8cm}<{\centering\arraybackslash}|
  p{0.95cm}<{\centering\arraybackslash}|
  p{0.95cm}<{\centering\arraybackslash}|
  p{0.55cm}<{\centering\arraybackslash}|}
  \hline
      Settings & 0 to 30 & 30 to 60 & 60 to 90 & All \\
    \hline
       Comparison 1 & 2.63 & 3.35 & 4.50 & 3.49\\ 
       Comparison 2 & 2.62 & 3.31 & 4.31 & 3.41\\ 
       Adopted setting & 2.66 & 3.30 & 4.27 & 3.41\\
    \hline
  \end{tabular}
  \vspace{-18pt}
\end{center}
\end{table}

\textbf{(3)} We further investigate the other two possible network designs of landmark-to-3DMM. Comparison 1 refines landmarks and regresses $\hat{\alpha}$ at the same step, and thus $\hat{\alpha}$ is regressed from $L^c$ in this setting. \textit{This is to study whether the reverse representation direction is benefited from refined landmarks $L^r$}. Comparison 2 further includes $z$, $\alpha_s$, and $\alpha_e$ in the landmark-to-3DMM regression in Fig. \ref{pipeline}, forming another multi-attribute feature aggregation to regress the $\hat{\alpha}$. \textit{This setting analyzes whether aggregation can also assist $\hat{\alpha}$ prediction}. 

From Table \ref{PGS_study}, results of Comparison 1 show that regressing $\hat{\alpha}$ from $L^c$ does not perform better than $L^r$ due to the finer and more accurate structure of $L^r$. Results of Comparison 2 show that multi-attribute aggregation used at the landmark-to-3DMM does not bring better performance. We assume this is because the information has been joined at the landmark refinement phase.

\textbf{Comparison to related work.} We benchmark performance on the widely-used AFLW2000-3D. The two versions of annotations (original and reannotated) are used. To have a fair comparison, we show results on the two different sets separately and compare them with other reported performances. Table \ref{FAL_AFLW2k_sota} shows the comparison on the original, and the reannotated version is in the supplementary. 
Our SynergyNet holds the best performance among all the related work on this standard dataset in Table \ref{FAL_AFLW2k_sota}. 
From the breakdown, our performance gain mainly comes from medium and large pose cases. We find that prior works encounter performance bottlenecks since they only regress 3DMM from images. However, referring to Table \ref{FAL_ab}, MAFA has already shown the best performance compared with prior arts due to its ability to fuse multi-attribute features. Then the landmark-to-3DMM further shows lower errors to break through the performance bottleneck.


\begin{table}[tb!]
\begin{center}
  \caption{\textbf{Benchmark on AFLW2000-3D for facial alignment.} The original annotation version is used. Our performance is the best with a gap over others on large poses.}
  \vspace{-7pt}
  \label{FAL_AFLW2k_sota}
  \footnotesize
  \begin{tabular}{|p{2.8cm}<{\centering}|
  p{0.85cm}<{\centering}|
  p{0.95cm}<{\centering}|
  p{0.95cm}<{\centering}|
  p{0.70cm}<{\centering}|}
  \hline
    AFLW2000-3D Original  &  0 to 30 & 30 to 60 & 60 to 90 & All \\
    \hline
    ESR \cite{cao2014face}& 4.60 & 6.70 & 12.67 & 7.99 \\
    3DDFA \cite{zhu2016face}& 3.43 & 4.24 & 7.17 & 4.94 \\
    Dense Corr \cite{yu2017learning}& 3.62 & 6.06 & 9.56 & 6.41 \\
    3DSTN \cite{bhagavatula2017faster} & 3.15 & 4.33 & 5.98 & 4.49\\
    3D-FAN \cite{bulat2017far} & 3.16 & 3.53 & 4.60 & 3.76 \\
    3DDFA-PAMI \cite{zhu2019face} & 2.84 & 3.57 & 4.96 & 3.79 \\
    PRNet \cite{feng2018joint} & 2.75 & 3.51 & 4.61 & 3.62 \\
    2DASL \cite{tu20203d} & 2.75 & 3.46 & 4.45 & 3.55 \\
    3DDFA-V2 (MR)\cite{guo2020towards} & 2.75 & 3.49 & 4.53 & 3.59 \\
    3DDFA-V2 (MRS)\cite{guo2020towards} & \textbf{2.63} & 3.42 & 4.48 & 3.51 \\
    SynergyNet (our) & \textbf{2.65} & \textbf{3.30} & \textbf{4.27} & \textbf{3.41} \\
    \hline
  \end{tabular}
  \vspace{-15pt}
\end{center}
\end{table}

Following 3DDFA-V2\cite{guo2020towards}, we next use the AFLW full set for evaluation (21K testing images with 21-point landmarks). We show the comparison in Table \ref{FAL_AFLW}. Our work has the best performance, especially with a performance gap over others on large-pose cases.

\begin{table}[tb!]
\begin{center}
  \caption{\textbf{Quantitative facial alignment comparison on AFLW with 21-point landmark definition.} }
  \label{FAL_AFLW}
  \footnotesize
  \begin{tabular}{|p{2.5cm}<{\centering}|
  p{0.85cm}<{\centering}|
  p{0.95cm}<{\centering}|
  p{0.95cm}<{\centering}|
  p{0.75cm}<{\centering}|}
  \hline
    AFLW  &  0 to 30 & 30 to 60 & 60 to 90 & All \\
    \hline
    ESR \cite{cao2014face} & 5.66 & 7.12 & 11.94 & 8.24 \\
    3DDFA \cite{zhu2016face} & 4.75 & 4.83 & 6.39 & 5.32 \\
    3D-FAN \cite{bulat2017far} & 4.40 & 4.52 & 5.17 & 4.69 \\
    3DSTN \cite{bhagavatula2017faster} & \textbf{3.55} & \textbf{3.92} & 5.21 & 4.23 \\
    3DDFA-PAMI \cite{zhu2019face} & 4.11 & 4.38 & 5.16 & 4.55 \\
    PRNet \cite{feng2018joint} & 4.19 & 4.69 & 5.45 & 4.77\\
    3DDFA-V2 \cite{guo2020towards} & 3.98 & 4.31 & 4.99 & 4.43 \\
    SynergyNet (our) & 3.76 & \textbf{3.92} & \textbf{4.48} & \textbf{4.06}\\
    \hline
  \end{tabular}
  \vspace{-11pt}
\end{center}
\end{table}

\subsection{Face Orientation Estimation Evaluation}

\textbf{Metrics and studies.} Following the evaluation protocol in \cite{yang2019fsa, ruiz2018fine,dairankpose, cao2021vector}, we calculate the mean absolute error (MAE) of predicted Euler angles in degrees. Groundtruth angles for each face in AFLW2000-3D are used, except for 31 samples whose yaw angles are outside the range [-99\degree, 99\degree]. We first study different combinations of the proposed modules in Table \ref{FOE_ab}. From the results, finer landmark structures from MAFA lead to better orientation estimation. The landmark-to-3DMM further contributes to better performance since pose parameter regression from 3D points leads to more robust poses than 2D images. 

We further study information fusion of using different attributes at MAFA and examine different parameter regression targets at the landmark-to-3DMM in Table \ref{FOE_LF_ab}. This analysis shows that more accurate orientation estimation is mainly benefited by pose regression at the reverse representation direction stage because Euler angles estimated from 3D representation are more robust than from 2D images. This breakdown also explains the performance gain of the landmark-to-3DMM in Table \ref{FOE_ab}.

\begin{table}[tb!]
\begin{center}
  \caption{\textbf{Ablative for face orientation estimation.} The same modules are studied in Table \ref{FAL_ab} for facial alignment. MAE of Euler angles in degree is reported.}
  \vspace{-7pt}
  \label{FOE_ab}
  \footnotesize
  \begin{tabular}[c]
  {|p{2.35cm}<{\centering\arraybackslash}|
  p{1.2cm}<{\centering\arraybackslash}|
  p{0.56cm}<{\centering\arraybackslash}|
  p{0.56cm}<{\centering\arraybackslash}|
  p{0.56cm}<{\centering\arraybackslash}|
  p{0.56cm}<{\centering\arraybackslash}|}
  \hline
  Multi-Attribute Feature Aggregation & Landmark-to-3DMM & Yaw & Pitch & Roll & Mean \\
    \hline
      - & - & 3.97 & 4.93 & 3.28 & 4.06\\ 
      \Checkmark & - & 3.72 & 4.37 & 2.88 & 3.65\\ 
      - & \Checkmark & 3.67 & 4.48 & 2.95 & 3.70\\ 
      \Checkmark & \Checkmark & \textbf{3.42} & \textbf{4.09} & \textbf{2.55} & \textbf{3.35}\\
    \hline
  \end{tabular}
  \vspace{-13pt}
\end{center}
\end{table}

\begin{table}[tb!]
\begin{center}
  \caption{\textbf{Study on different attributes and different regression targets for face orientation estimation.} The same structures are also studied in Table \ref{FAL_PIFV_ab} for the alignment task. The first three rows study different attributes for aggregation. Based on the landmark refinement, Row 4 to 6 further study performance gains of different regression targets at the reverse representation stage.}
  \vspace{-7pt}
  \footnotesize
  \label{FOE_LF_ab}
  \begin{tabular}[c]
  {|
  p{3.8cm}<{\arraybackslash}|
  p{0.65cm}<{\centering\arraybackslash}|
  p{0.65cm}<{\centering\arraybackslash}|
  p{0.65cm}<{\centering\arraybackslash}|
  p{0.6cm}<{\centering\arraybackslash}|}
  \hline
      Structures  & Yaw & Pitch & Roll & Mean \\
    \hline
       Point-feature only & 3.81 & 4.42 & 2.89 & 3.71\\
       Point + image feature & 3.72 & 4.39 & 2.85 & 3.66\\ 
       MAFA & 3.72 & 4.37 & 2.88 & 3.65\\
       MAFA+ L$\rightarrow$3D(pose) &  3.58 & 4.06 & 2.57 & 3.40\\ 
       MAFA+ L$\rightarrow$3D(shape,expr) & 3.47 & 4.23 & 2.59 & 3.43\\
       MAFA+ L$\rightarrow$3D(all) &  3.42 & 4.09 & 2.55 & 3.35\\ 
    \hline
  \end{tabular}
  \vspace{-10pt}
\end{center}
\end{table}

\textbf{Benchmark comparison.} We collect works that focus only on face orientation estimation \cite{ruiz2018fine, yang2019fsa, hsu2018quatnet, dairankpose, cao2021vector} and 3DMM-based methods \cite{zhu2016face, zhu2019face, guo2020towards, tu20203d}. The 3DMM-based works do not include evaluation of this task. To show the full benchmark list, we evaluate their methods using the released pretrained models. Table \ref{FOE_sota} shows that our method is the best and holds a performance gap over others. We display visual comparison in Fig. \ref{compare_FAL_FOE} and more studies and discussion in the supplementary.

\begin{figure}[tb!]
    \centering
    \vspace{-8pt}
    \includegraphics[width=1.0\linewidth]{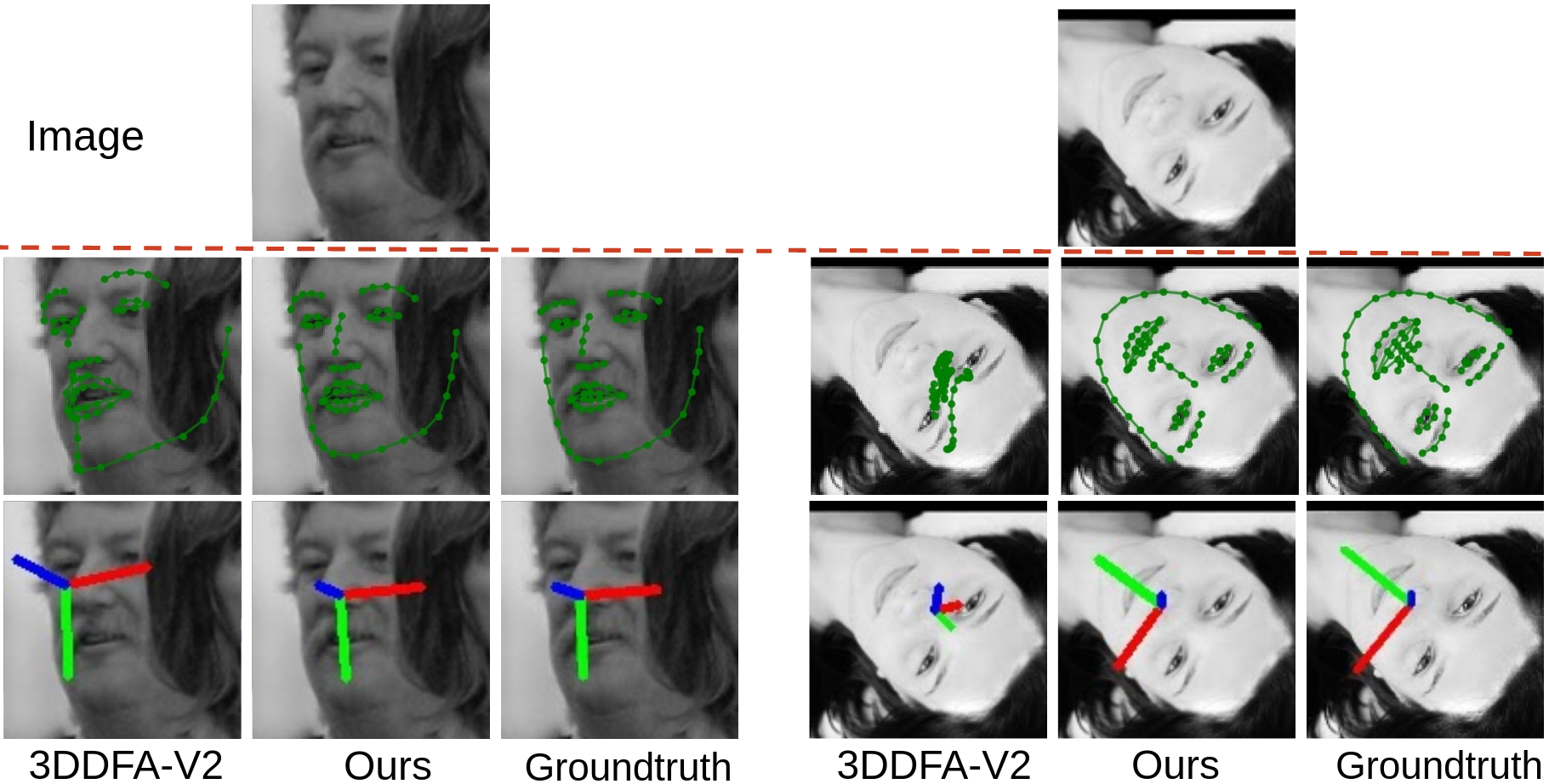}
    \caption{\textbf{Qualitative comparison of facial alignment and orientation estimation.} The case on the left is low-resolution, blurry, and thus challenging. The case on the right is of rare and extreme roll rotation. Our results show more robustness over 3DDFA-V2.}
    \label{compare_FAL_FOE}
    \vspace{-2pt}
\end{figure}

\begin{table}[tb!]
\begin{center}
  \caption{\textbf{Face orientation estimation benchmark comparison on ALFW2000-3D.} PnP shows solving perspective-n-point problems using groundtruth landmarks. We do not include PRNet here since it does not infer face orientation directly and also obtains poses by PnP with predicted landmarks.}
  \vspace{-7pt}
  \label{FOE_sota}
  \footnotesize
  \begin{tabular}{|p{3.0cm}<{\centering}|
  p{0.8cm}<{\centering}|
  p{0.8cm}<{\centering}|
  p{0.8cm}<{\centering}|
  p{0.75cm}<{\centering}|}
  \hline
    AFLW2000-3D  & Yaw & Pitch & Roll & Mean \\
    \hline
    PnP-landmark & 5.92 & 11.76 & 8.27 & 8.65 \\
    FAN-12 point \cite{bulat2017far}& 6.36 & 12.30 & 8.71 & 9.12 \\
    HopeNet \cite{ruiz2018fine}& 6.47 & 6.56 & 5.44 & 6.16 \\
    SSRNet-MD \cite{ssrnet}& 5.14 & 7.09 & 5.89 & 6.01 \\
    FSANet \cite{yang2019fsa}& 4.50 & 6.08 & 4.64 & 5.07 \\
    QuatNet \cite{hsu2018quatnet}& 3.97 & 5.62 & 3.92 & 4.15\\
    TriNet \cite{cao2021vector} & 4.20 & 5.77 & 4.04 & 3.97 \\
    RankPose \cite{dairankpose}& \textbf{2.99} & 4.75 & 3.25 & 3.66 \\
    3DDFA-TPAMI \cite{zhu2019face} & 4.33 & 5.98 & 4.30 & 4.87 \\
    2DASL \cite{tu20203d} & 3.85 & 5.06 & 3.50 & 4.13\\
    3DDFA-V2 \cite{guo2020towards} & 4.06 & 5.26 & 3.48 & 4.27 \\
    SynergyNet (our) & 3.42 & \textbf{4.09} & \textbf{2.55} & \textbf{3.35} \\
    \hline
  \end{tabular}
  \vspace{-8pt}
\end{center}
\end{table}

\subsection{3D Face Modeling Evaluation}
\label{3dface_recon}
\begin{figure}[tb!]
    \centering
    \vspace{-3pt}
    \includegraphics[width=1.0\linewidth]{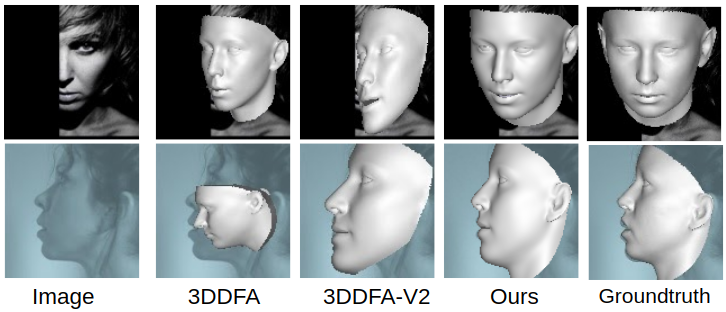}
    \caption{\textbf{Qualitative comparison of 3D face models.} Our results are robust to rare and out-of-domain face examples.}
    \label{Qual_3DFace}
    \vspace{-5pt}
\end{figure}
\textbf{Metrics and Comparison}. Following \cite{feng2018joint, tu20203d, guo2020towards}, we first evaluate 3D face modeling on AFLW2000-3D. Two protocols are used. Protocol 1 suggested by \cite{feng2018joint, tu20203d, zhu2016face} uses the iterative closet point (ICP) algorithm to register groundtruth 3D models and predicted models. NME of per-point error normalized by interocular distances is calculated. Protocol 2, suggested by \cite{guo2020towards} and also called dense alignment, calculates the per-point error normalized by bounding box sizes with groundtruth models aligned with images. Since ICP is not used, pose estimation would affect the performance under this protocol, and the NME would be higher. We illustrate numerical comparison in Table \ref{ALFW2K_Recon}. The results show the ability of SynergyNet to recover 3D face models from monocular inputs and attain the best performance. In addition, we further exhibit visual comparison in Fig. \ref{Qual_3DFace}. Our SynergyNet is capable of recovering 3D faces under rare and out-of-domain scenarios, such as heavily cropped or underwater cases.

\begin{table}[tb!]
\begin{center}
  \caption{\textbf{3D face modeling comparison on AFLW2000-3D.} Refer to Sec. \ref{3dface_recon} for the protocol details.}
  \label{ALFW2K_Recon}
  \footnotesize
  \begin{tabular}{|p{1.3cm}<{\centering}|
  p{0.8cm}<{\centering}|
  p{0.8cm}<{\centering}|
  p{0.7cm}<{\centering}|
  p{0.8cm}<{\centering}|
  p{1.1cm}<{\centering}|}
    \hline
     Protocol-1 \cite{feng2018joint, tu20203d, zhu2016face}& 3DDFA\cite{zhu2016face} & DeFA\cite{liu2017dense} & PRNet\cite{feng2018joint} & 2DASL\cite{tu20203d} & SynergyNet (our) \\
    \cline{1-6}
    NME & 5.37 & 5.55 & 3.96 & 2.10 & \textbf{1.97}\\
    \hline
  \end{tabular}
  \begin{tabular}{|p{1.5cm}<{\centering}|
  p{0.8cm}<{\centering}|
  p{0.8cm}<{\centering}|
  p{1.6cm}<{\centering}|
  p{1.3cm}<{\centering}|}
    \hline
    Protocol-2 \hspace{1cm}\cite{guo2020towards} & 3DDFA \cite{zhu2016face} & DeFA\cite{liu2017dense} & 3DDFA-V2 \hspace{0.5cm}\cite{guo2020towards} & SynergyNet (our) \\
     \cline{1-5}
    NME & 6.56 & 6.04 & 4.18 & \textbf{4.06}\\
    \hline
  \end{tabular}
  \vspace{-13pt}
\end{center}
\end{table}

Next, we evaluate the performance of 3D face modeling on Florence \cite{florence2011} with real scanned 3D faces. We follow the protocol from \cite{guo2020towards, feng2018joint}, which renders 3D face models on different views with pitch of -15, 20, and 25 degrees and yaw of -80, -40, 0, 40, and 80 degrees. The rendered images are used as the test inputs. After reconstruction, face models are cropped to 95mm from the nose tip, and ICP is performed to calculate point-to-plane root mean square error (RMSE) with cropped groundtruth. Numerical results are shown in Table \ref{Florence_sota}. An error curve that shows our robustness to yaw angle changes and a qualitative comparison on Florence are displayed in the supplementary. 
\begin{table}[tb!]
\begin{center}
  \caption{\textbf{3D face modeling comparison on Florence.} Point-to-plane RMSE is calculated for evaluation.}
  \vspace{-7pt}
  \footnotesize
  \label{Florence_sota}
  \begin{tabular}{|p{1.2cm}<{\centering}|
  p{0.8cm}<{\centering}|
  p{1.0cm}<{\centering}|
  p{1.7cm}<{\centering}|
  p{1.5cm}<{\centering}|}
  \hline
    Florence & PRNet\cite{feng2018joint} & 2DASL\cite{tu20203d} & 3DDFA-V2 \quad\quad \cite{guo2020towards} & SynergyNet\quad(our) \\
    \hline
    RMSE & 2.25 & 2.05 & 2.04 & \textbf{1.87} \\
    \hline
  \end{tabular}
  \vspace{-20pt}
\end{center}
\end{table}

\section{Conclusion}
\vspace{-4pt}
This work proposes a synergy process that utilizes the relation between 3D landmarks and 3DMM parameters, and they collaboratively contribute to better performance. We establish a representation cycle, including forward direction, from 3DMM to 3D landmarks, and reverse representation direction, from 3D landmarks to 3DMM. Specifically, We propose two modules, multi-attribute feature aggregation for landmark refinement and the landmark-to-3DMM module. Extensive experiments validate our network design, and we show a detailed performance breakdown for each included attribute and regression target. Our SynergyNet only adopts simple network operations and attains superior performance, making it a fast, accurate, and easy-to-implement method.


\section*{Acknowledgement}
We sincerely thank Jingjing Zheng, Jim Thomas, and Cheng-Hao Kuo for their detailed feedback on this paper.

{\small
\bibliographystyle{ieee_fullname}
\bibliography{egbib}
}

\clearpage
\newpage
\pagebreak
\begin{center}
\textbf{\large Supplemental Materials:}
\end{center}
\setcounter{section}{0}
\setcounter{equation}{0}
\setcounter{figure}{0}
\setcounter{table}{0}
\setcounter{page}{1}
\makeatletter
\renewcommand{\theequation}{S\arabic{equation}}
\renewcommand{\thefigure}{S\arabic{figure}}
\renewcommand{\thetable}{S\arabic{figure}}
\renewcommand\thesection{\Alph{section}}
\renewcommand\thesubsection{\thesection.\Alph{subsection}}

\section{Overview}

We document this supplementary into the following sections. In Section \ref{c}, we provide details of our network architectures and loss weights for training. We further present a study for network backbone choices and a study for showing that our performance gain is not simply from using more network parameters. In Section \ref{e}, evaluation of facial alignment on AFLW2000-3D reannotation version is exhibited. In Section \ref{g}, we present analysis and discussion on the reverse representation direction. In Section \ref{f}, we dig into performance comparison using $L^r$ and $L^c$. In Section \ref{h}, an error curve and visual comparison of 3D face modeling on Florence are illustrated. In Section \ref{b}, we describe texture synthesis using introduced UV-texture GAN and further compare with textures from 3DMM fitting. In Section \ref{i}, we add more qualitative results from our face geometry prediction using the 300VW video dataset and Artistic Faces.

\begin{figure*}[hbt]
    \centering
    \includegraphics[width=1.0\linewidth]{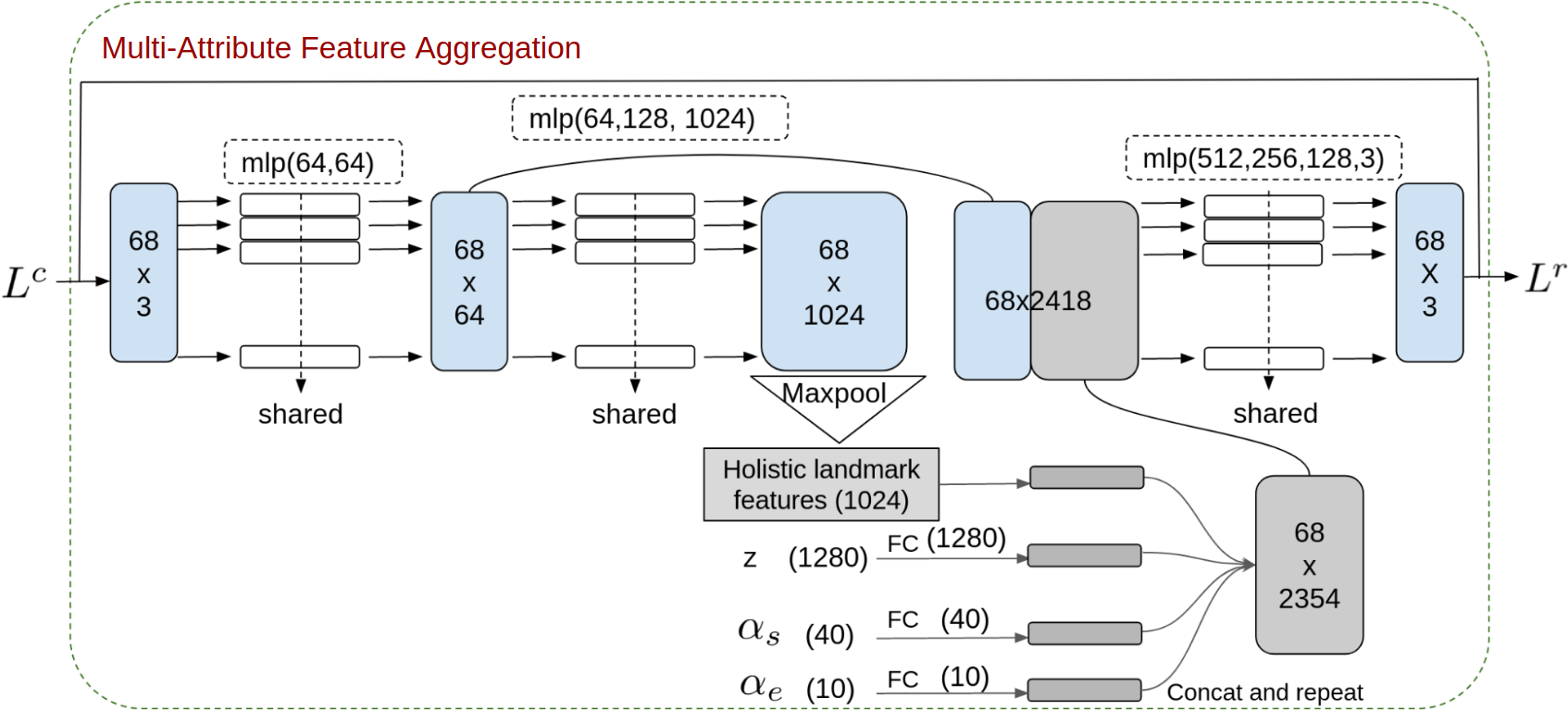}
    \caption{\textbf{Detailed structure of MAFA.} mlp(64,64) means two MLP layers with output channel sizes 64 and 64. ReLU and batch normalization are used for each layer. The notations correspond to those in the main paper. ($z$ is latent image feature, $\alpha_s$ and $\alpha_e$ are 3DMM shape and expression parameters regressed from images, $L^c$ and $L^r$ are 3D landmarks before and after the landmark refinement.) }
    \label{m2fa}
    \vspace{-5pt}
\end{figure*}

\begin{figure}[hbt]
    \centering
    \includegraphics[width=1.0\linewidth]{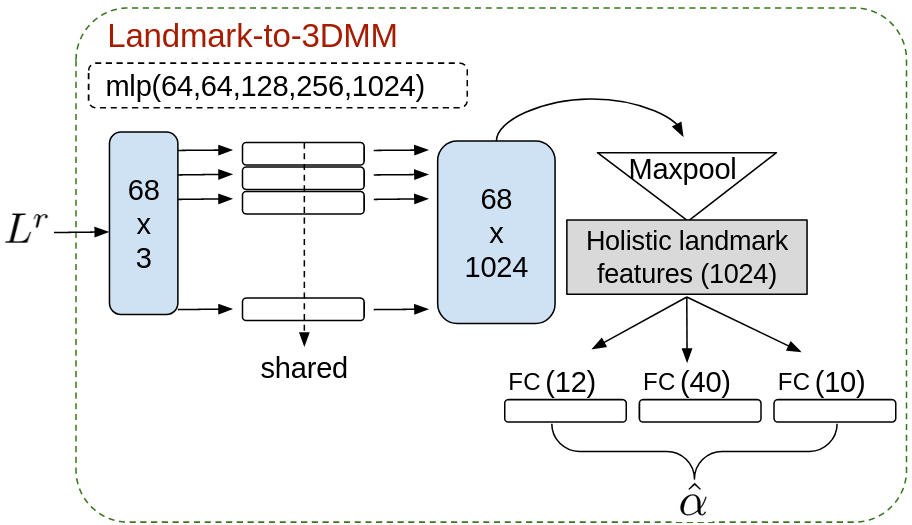}
    \caption{\textbf{Detailed structure of the landmark-to-3DMM module.} The notations correspond to those in the main paper. $L^r$ is refined 3D landmarks, and $\hat{\alpha}$ is regressed landmark geometry.}
    \label{lgs}
    \vspace{-5pt}
\end{figure}

\section{Network Architecture, Hyper-Parameters, and Network Parameter Studies}
\label{c}

\textbf{Details of architecture.} Based on Fig. 2 in the main paper (pipeline graph of our SynergyNet), detailed network architecture is described here. Following \cite{guo2020towards}, we use MobileNet-V2 as the backbone for 3DMM regression from images. The latent image features $z$ after the global max-pooling is 1280-dim. The pose, shape, and expression decoders are fully-connected (FC) layers with input $z$ and output 3DMM parameters of 12 ($\alpha_p$), 40 ($\alpha_s$), and 10 ($\alpha_e$) dimensions for pose, shape, and expression.

Fig. \ref{m2fa} shows the network architecture of multi-attribute feature aggregation.
Aggregation of the latent image features, $\alpha_s$, and $\alpha_e$ and global point features forms a 2354-dim feature vector. We repeat this vector and append it to the low-level point features to obtain multi-attribute point features, whose size is 68$\times$2418. Later we use another MLP-block to obtain refined landmarks $L^r$. 

Fig. \ref{lgs} illustrates the architecture of the landmark-to-3DMM module. With $L^r$ as the module input, this module reverses the representation direction and regresses 3DMM parameters $\hat{\alpha}$, also referred to as landmark geometry in the paper.

\textbf{Loss weights.} For weights of loss terms (Eq.7 in paper), we choose $\lambda_1 = 0.02$, $\lambda_2 = 0.03$, $\lambda_3 = 0.02$, and $\lambda_4 = 0.001$ for training.

\textbf{Study on the backbone for 3DMM regression from images.} We next conduct a study on the backbone choices for facial alignment using the AFLW full set. We select MobileNet-V2 \cite{sandler2018mobilenetv2}, ResNet50 \cite{he2016deep}, ResNet101 \cite{he2016deep}, ResNeSt50 \cite{zhang2020resnest}, and  ResNeSt101\cite{zhang2020resnest} for comparison. From Table \ref{FAL_AFLW_backbone}, one could see that the 101-residual layer network is too deep, and thus the performance drops compared with the 50-residual layer network. ResNeSt, a split-attention variant of ResNet, can improve the performance for the 101-residual layer case, but the performance of ResNeSt50 is on par with ResNet50. We think this is because the split-attention scales better and remedies the undesirable effects of deeper networks, which are described in their work \cite{zhang2020resnest}.

\begin{table}[tb!]
\begin{center}
  \caption{\textbf{Backbone study on the AFLW full set.} We compare MobileNet \cite{sandler2018mobilenetv2}, ResNet \cite{he2016deep}, and ResNeSt \cite{zhang2020resnest}, a ResNet variant with split-attention.}
  \vspace{-10pt}
  \label{FAL_AFLW_backbone}
  \begin{tabular}{|p{2.0cm}<{\centering}|
  p{1.0cm}<{\centering}|
  p{1.2cm}<{\centering}|
  p{1.2cm}<{\centering}|
  p{0.75cm}<{\centering}|}
  \hline
    Backbone  &  0 to 30 & 30 to 60 & 60 to 90 & All \\
    \hline
    MobileNet & 3.86 & 4.13 & 4.61 & 4.20 \\
    ResNet50 & 3.76 & 3.92 & 4.48 & 4.06 \\
    ResNeSt50 & 3.76 & 3.92 & 4.52 & 4.07 \\
    ResNet101 & 3.90 & 4.14 & 5.08 & 4.38\\
    ResNeSt101 & 3.78 & 4.04 & 4.62 & 4.15\\
    \hline
  \end{tabular}
  \vspace{-15pt}
\end{center}
\end{table}

\textbf{Study on the number of network parameters.} We further conduct a study to verify the effectiveness of the introduced multi-attribute feature aggregation (MAFA) for landmark refinement and the landmark-to-3DMM modules. The following experiment shows that our performance gain comes from designing the two proposed modules in our synergy process, rather than simply using more network parameters.

By using MobileNet-V2 as the face image encoder backbone, network parameters of our SynergyNet amount to 3.8M (3.0M for the backbone) and 0.8M for the MAFA and landmark-to-3DMM modules). We build another baseline model, Image $\to$ 3DMM (larger), that only contains 3DMM parameter regression from images using more network parameters. We add additional MLP layers with ReLU and BN, which amount to 0.8M parameters, after the image bottleneck feature $z$ for regressing $\alpha_p$, $\alpha_s$, and $\alpha_e$. Thus, this baseline model and our SynergyNet have approximately the same number of network parameters. 

In Table \ref{network_params}, we show experiments with the baseline model on facial alignment and face orientation estimation. More network parameter adoption for regressing 3DMM from images only leads to minor improvements. Especially for facial alignment, using extra 0.8M parameters of MLPs only gives 0.02 overall performance gain. The results validate the designed synergy process. Without the proposed modules, using more parameters only brings minor improvements.

\begin{table*}[htb!]
\begin{center}
  \caption{\textbf{Comparison with the baseline that simply uses more network parameters.} The first table shows results on AFLW2000-3D Original for facial alignment, and the second table shows results also on AFLW2000-3D for face orientation estimation. \# of params means the number of network parameters. More network parameter use for the baseline 3DMM regression from images only results in a limited performance gain. The experiment validates that the performance gain of our SynergyNet is not simply from more parameter adoption.}
  \label{network_params}
  \begin{tabular}{|p{3.7cm}<{}|
  p{2.0cm}<{\centering}|
  p{1.2cm}<{\centering}|
  p{1.2cm}<{\centering}|
  p{1.2cm}<{\centering}|
  p{0.75cm}<{\centering}|}
  \hline
    Structures  & \# of params & 0 to 30 & 30 to 60 & 60 to 90 & All \\
    \hline
    Image $\to$ 3DMM & 3.0M & 2.99 & 3.80 & 4.86 & 3.88 \\
    Image $\to$ 3DMM (larger) & 3.8M & 2.98 & 3.82 & 4.77 & 3.86 \\
    SynergyNet & 3.8M & 2.66 & 3.30 & 4.27 & \textbf{3.41} \\
    \hline
  \hline
      Structures & \# of params & Yaw & Pitch & Roll & Mean \\
    \hline
       Image $\to$ 3DMM & 3.0M & 3.97 & 4.93 & 3.28 & 4.06\\
       Image $\to$ 3DMM (larger) & 3.8M & 3.80 & 4.62 & 2.84 & 3.75\\ 
       SynergyNet & 3.8M & 3.42 & 4.09 & 2.55 & \textbf{3.35}\\
    \hline
  \end{tabular}
  \vspace{-15pt}
\end{center}
\end{table*}

\section{Evaluation on Reannotated AFLW2000-3D}
\label{e}
Reannotation of AFLW2000-3D is provided in LS3D-W\cite{bulat2017far}. Few works, Deng \textit{et al} \cite{deng2019accurate}, DHM \cite{sun2018deep} and MCG-Net \cite{shang2020self}, report their performance on the annotated version. To aggregate more results for this study, we also include evaluation of 3DDFA \cite{zhu2016face}, PRNet \cite{feng2018joint}, and 3DDFA-V2 \cite{guo2020towards} using their pretrained models. From Table \ref{FAL_reanno}, normalized mean errors (NMEs) are generally lower than using the original annotation. This shows the higher quality of the reannotation. Among the methods for comparison, our result is the best and holds a performance gap over others. Compared with PRNet, the second-best method in the table, our improvements are derived from large pose cases. 

\begin{table}[tb!]
\begin{center}
  \caption{\textbf{Comparison on AFLW2000-3D Reannotation.} Our method has the best alignment result and holds a performance gap over others.}
  \label{FAL_reanno}
  \vspace{-5pt}
  \begin{tabular}{|p{2.49cm}<{\centering}|
  p{1.0cm}<{\centering}|
  p{1.2cm}<{\centering}|
  p{1.2cm}<{\centering}|
  p{0.65cm}<{\centering}|}
  \hline
    AFLW2000-3D Reannotated  &  0 to 30 & 30 to 60 & 60 to 90 & All \\
    \hline
    DHM \cite{sun2018deep} & 2.28 & 3.10 & 6.95 & 4.11 \\
    {3DDFA \cite{zhu2016face}} & 2.84 & 3.52 & 5.15 & 3.83 \\
    PRNet \cite{feng2018joint} & 2.35 & 2.78 & 4.22 & 3.11 \\
    MGCNet \cite{shang2020self} & 2.72 & 3.12 & 3.76 & 3.20 \\
    Deng \textit{et al} \cite{deng2019accurate} & 2.56 & 3.11 & 4.45 & 3.37 \\
    3DDFA-V2 \cite{guo2020towards} & 2.84 & 3.03 & 4.13 & 3.33 \\
    SynergyNet (our) & \textbf{2.05} & \textbf{2.49} & \textbf{3.52} & \textbf{2.65} \\
    \hline
  \end{tabular}
  \vspace{-18pt}
\end{center}
\end{table}

\section{Discussion on Reverse Representation Direction}
\label{g}
\textbf{Why use sparse landmarks rather than full vertices?} Landmark geometry $\hat{\alpha}$ in Sec.3.3 of the main paper describes revealing facial geometry underlying in sparse 3D landmarks. In contrast to sparse landmarks (68 points in our work), mesh from BFM Face includes 53.5K vertices (45K if excluding the neck and ears). When surveying on point processing, research usually adopts only 1024 or 2048 points \cite{qi2017pointnet, qi2017pointnet++, xu2020grid}. Much denser points are inefficient for point processing, and the accommodation is also limited by GPU memory. On the other hand, because facial alignment is considered as an upstream task for the downstream application such as face recognition \cite{shi2006effective} or recent streaming video compression \cite{NvidiaMaxine, wang2020one}, high efficiency is more desirable.

Although a point-sampling strategy could be used for downsizing, 3D landmarks are very efficient and compact for expressing facial traits and outlines. Therefore, 3D landmarks are desirable for predicting facial geometry, and our focus of this work is to exert the synergy between facial landmarks and 3DMM parameters, which collaboratively contribute to better performance.

\textbf{Advantage of the reverse representation direction.} Compared with 2D images, 3D sparse landmarks describe facial traits and approximate face outlines. Although landmarks are sparse, the representation provides another view to learn facial geometry and complements with 3DMM regressed from images. For example, facial geometry for large pose cases is hard to estimate from the 2D due to self-occlusion. Further, the face orientation is defined in the 3D space; thus, it is more advantageous to estimate face orientation from 3D points, whose learning paradigm provides less ambiguity. From Table 2 and 7 in the main paper that study contributions for each regression target at the $L\to$ 3D stage, MAFA+$L\to$ 3D(all) improves the performance on facial alignment and face orientation estimation. The results show the ability of the reverse representation direction.

We also conduct evaluations using $\hat{\alpha}$ as the output on facial alignment and face orientation estimation using AFLW2000-3D. The 3D landmarks reconstructed by $\hat{\alpha}$ and the face orientation converted from its pose parameter $\hat{\alpha_p}$ attain an NME of 6.48 on the alignment and an MAE of 5.76 on the orientation estimation. The results are reasonable since the direct input to the landmark-to-3DMM module is $L^r$, 68-point sparse landmarks that present only approximate facial traits and outlines. However, these numerical results are comparable with some methods in Table 4 and 8 of the main paper. The results validate our training strategy so that \textit{3DMM estimation directly from sparse landmarks achieves on par performance with some studies for 3DMM regression from images}, whose information lies on dense grids.

\section{Evaluation on $L^r$ v.s. $L^c$} 
\label{f}
From Table 1 and 6 in the paper (1$^{st}$ row: without refinement and use $L^c$ for evaluation; 2$^{nd}$ row: with refinement and use $L^r$ for evaluation), one can observe from these two tables that with the refinement branch, the performance is significantly improved. Specifically, Table 1-facial alignment error: 3.88 ($L^c$) to 3.49 ($L^r$); Table 6-face orientation error: 4.06 ($L^c$) to 3.65 ($L^r$). 

\begin{figure}[bt!]
\begin{center}
\includegraphics[width=1.0\linewidth]{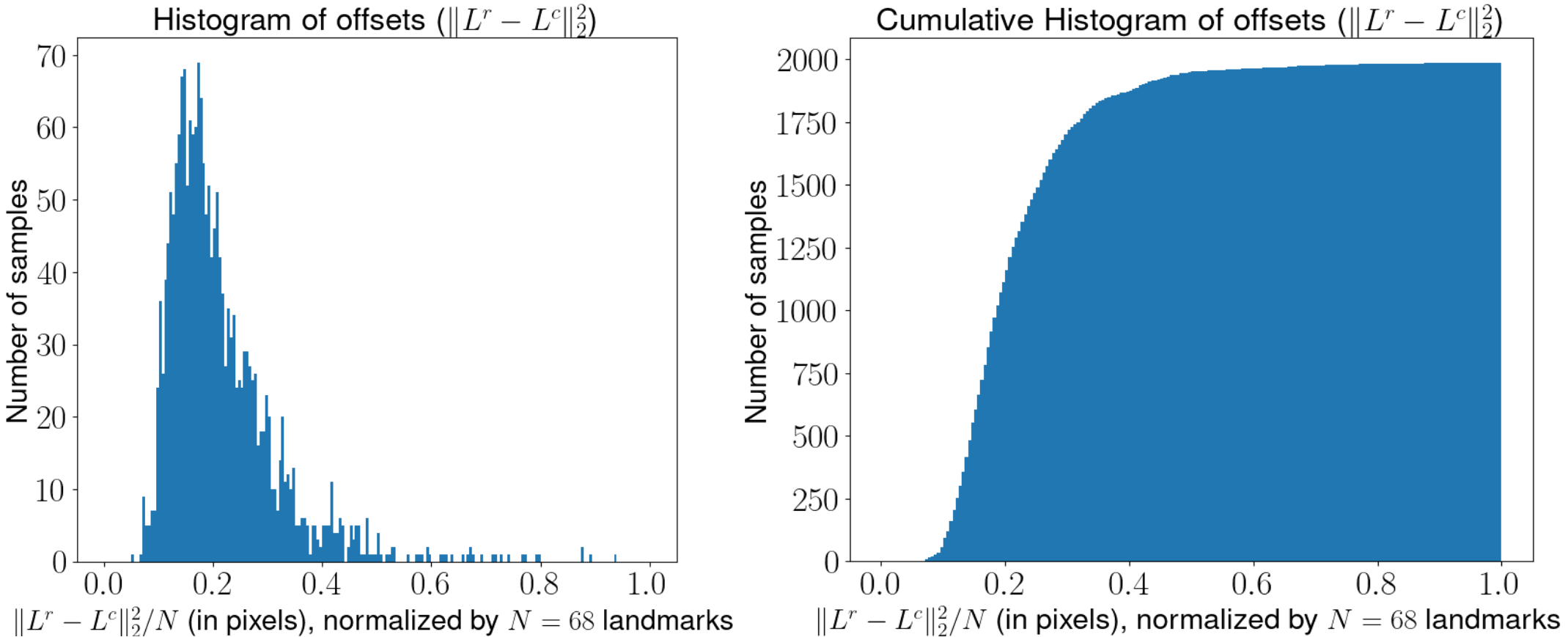}
\end{center}
  \vspace{-5pt}
  \caption{Histogram and cumulative histogram of the offset term: $\|L^r-L^c\|^2_2$. These plots show the difference of the landmark set before and after refinement.}
\label{f2}
\vspace{-5pt}
\end{figure}

\begin{figure}[bt!]
\begin{center}
\includegraphics[width=0.97\linewidth]{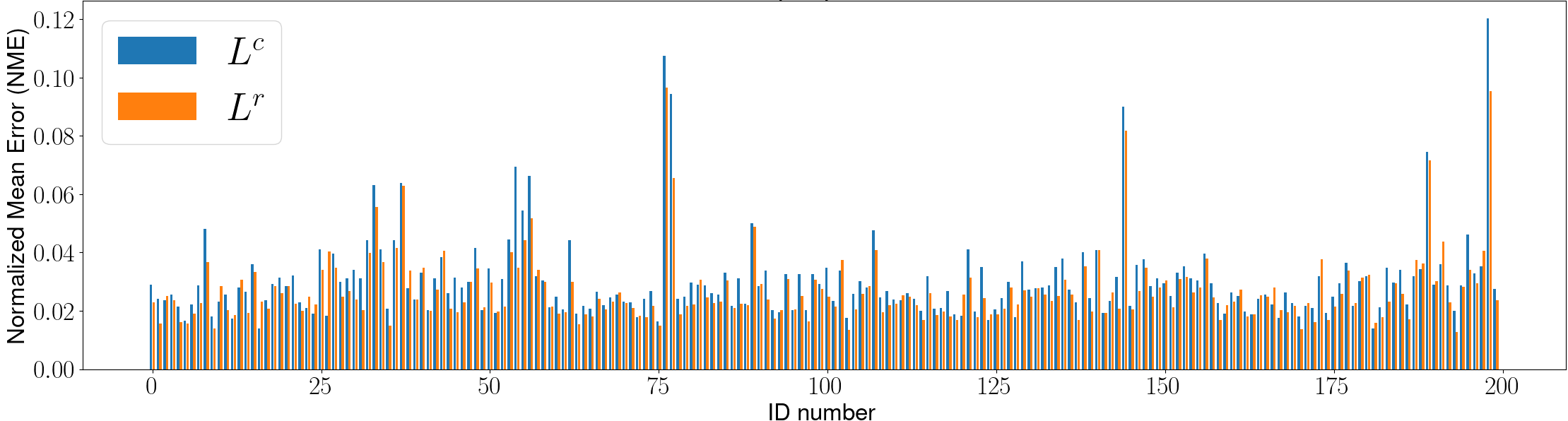}
\end{center}
  \vspace{-5pt}
  \caption{NME of random 200 people: The lower the better. Zoom in for the best view.}
\label{f3}
\vspace{-5pt}
\end{figure}

We further show the histograms of offsets $\|L^r-L^c\|^2_2$ in Fig.\ref{f2}. We use AFLW2000-3D for evaluation. One can see that the difference of $L^r$ and $L^c$ peaks at 0.22 pixels for each landmark. This matches the purpose of refinement that by predicting each landmark more precisely, the total improvements can break through the performance bottleneck in the benchmark list (paper Table 4). In addition, we plot the normalized mean error (NME) for random 200 people in Fig.\ref{f3} (zoom in for the best view). One can find that the blue bars are higher overall, meaning the error using $L^c$ is higher than $L^r$.

\section{3D Faces on Florence}
\label{h}
\begin{figure}[bt!]
    \centering
    \includegraphics[width=1.0\linewidth]{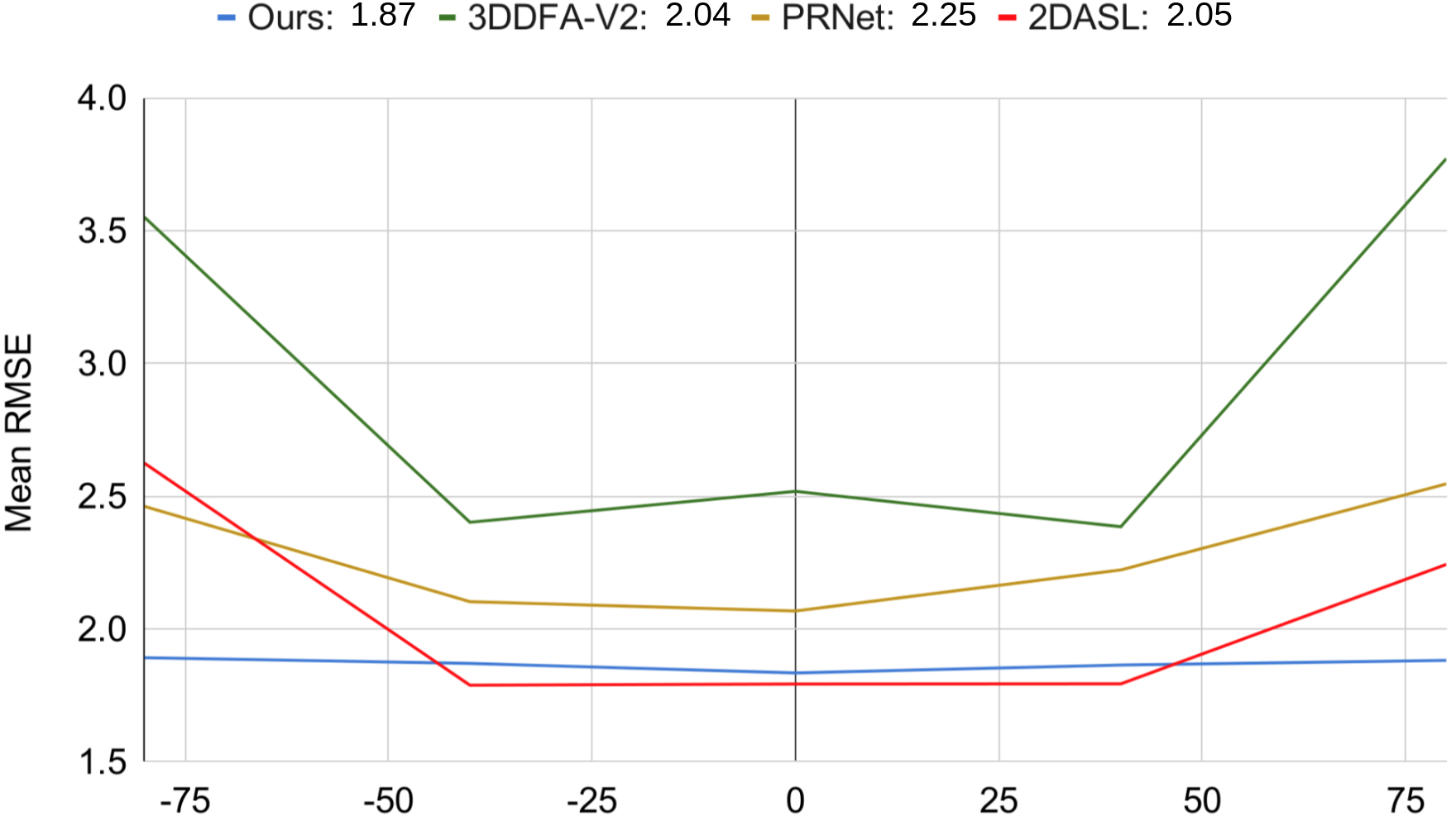}
    \caption{\textbf{Error curve for 3D face modeling by yaw angle on the Florence dataset.} The numbers at the top are mean RMSE over all testing data. Our method is rather robust to pose changes and attains the lowest overall point-to-plane RMSE.}
    \label{florence}
    \vspace{-1pt}
\end{figure}

Based on the Florence experiment in Section 4.3 and Table 10 of the main paper, we further show an error curve comparison in Fig. \ref{florence} for 3D face modeling. Our method is robust to pose changes and attains a \textit{nearly flat} error curve. Although the results of 2DASL for low and medium cases are close to ours, they are not robust for large pose cases.

We visualize the reconstructed meshes and compare with the current best-performing 3DMM-based method (3DDFA-V2 \cite{guo2020towards}) and UV-position-based method (PRNet \cite{feng2018joint}) in Fig. \ref{florence_recon_comp}. We mark point-to-plane RMSEs beside each face model. Our reconstructed faces show narrower eye-to-side distances, higher cheeks, and pointed chins from the upper example. These features are consistent with the groundtruth model. On the other hand, 3DDFA-V2 shows wider faces, unapparent cheeks, and non-pointed chins; thus, their errors are higher. Besides, for a cropping range of 95mm from the nose tip, 3DDFA-V2 shows more forehead areas than the groundtruth model, which means the geometry prediction is inaccurate. PRNet is not 3DMM-based. Although PRNet has higher flexibility to predict per-vertex deformation due to its non-parametric nature, it is also harder to estimate a precise 3D face via vertex regression on a UV-position map. From the lower example, our faces are wider and consistent with the groundtruth. In contrast, 3DDFA-V2 shows more elongated shapes for large pose cases. PRNet also shows skewed faces under large pose scenarios.

\section{Texture Synthesis}
\label{b}
\begin{figure}[bt!]
    \centering
    \includegraphics[width=1.0\linewidth]{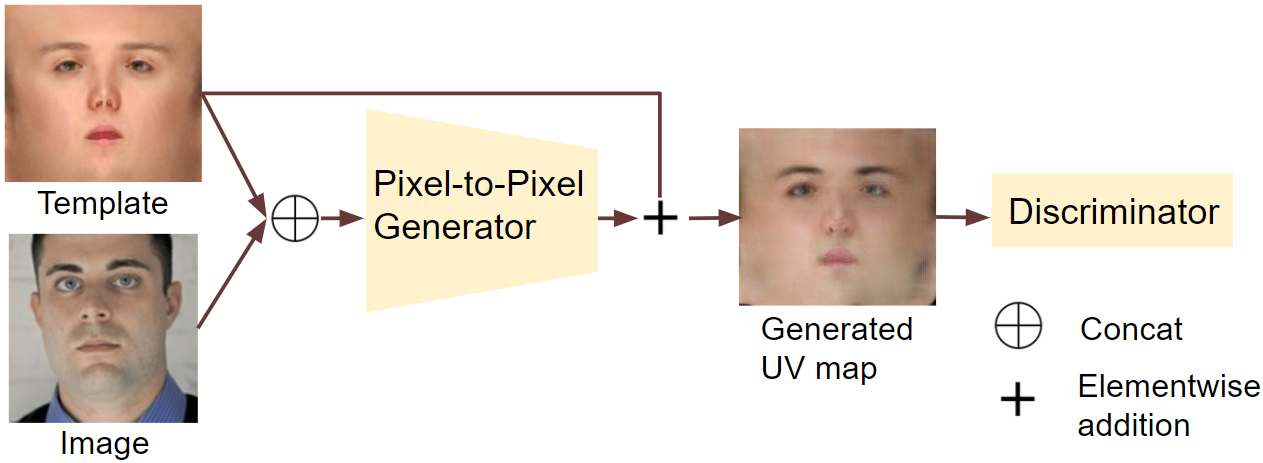}
    \caption{\textbf{UV-texture GAN.} The generator produces a UV map from a fixed template and an input image. The generated UV map combines structures of the template and the skin color of the image.}
    \label{uvgan}
    \vspace{-3pt}
\end{figure}

Most previous works for 3D facial alignment via 3D face modeling mainly focus on the geometry \cite{zhu2016face, zhu2019face, guo2020towards, feng2018joint, tu20203d, albiero2021img2pose}. To get more realistic 3D face models, here we also conduct a smaller study on texture synthesis based on our predicted 3D face models.

Similar to 3DMM fitting for 3D faces, as illustrated in the main paper Eq.(1), textures can also be synthesized by adding a mean texture term and a multiplication term of texture basis and parameters. For example, BFM Face contains texture parameters with a 199-dim texture basis. However, 3DMM texture fitting usually produces over-smooth textures that lack reality. (See examples in Fig. \ref{uv_compare}). 

Here we introduce a simple but effective UV-texture Generative Adversarial Network (UV-texture GAN) for texture synthesis. The model structure is illustrated in Fig. \ref{uvgan}. UV mapping \cite{foley1996computer} involves per-vertex color mappings from UV-texture maps. Each vertex is associated with its $(u,v)$-coordinate for querying vertex color from the three-channel UV-texture maps.

\begin{figure}[bt!]
    \centering
    \includegraphics[width=0.90\linewidth]{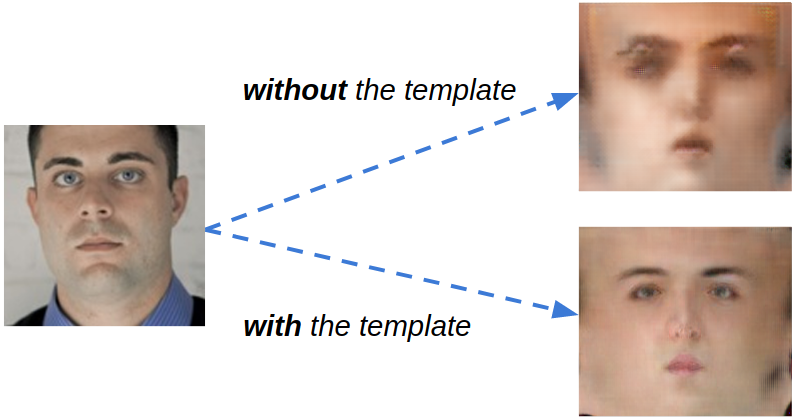}
    \caption{\textbf{Effects of using the template in the UV-texture GAN}. Without the template shown in Fig. \ref{uvgan}, facial traits such as eyes and moth are blurry and inaccurate.}
    \label{template_compare}
    \vspace{-4pt}
\end{figure}

\begin{figure}[bt!]
    \centering
    \includegraphics[width=1.0\linewidth]{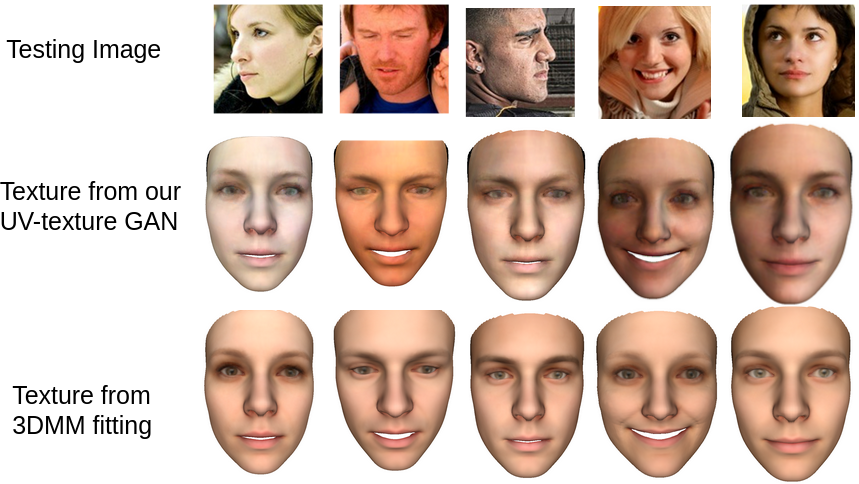}
    \caption{\textbf{Synthesized texture comparison.} Textures synthesized by the introduced UV-texture GAN are more realistic than textures from 3DMM fitting.}
    \label{uv_compare}
    \vspace{-5pt}
\end{figure}

\begin{figure*}[bt!]
    \centering
    \includegraphics[width=0.85\linewidth]{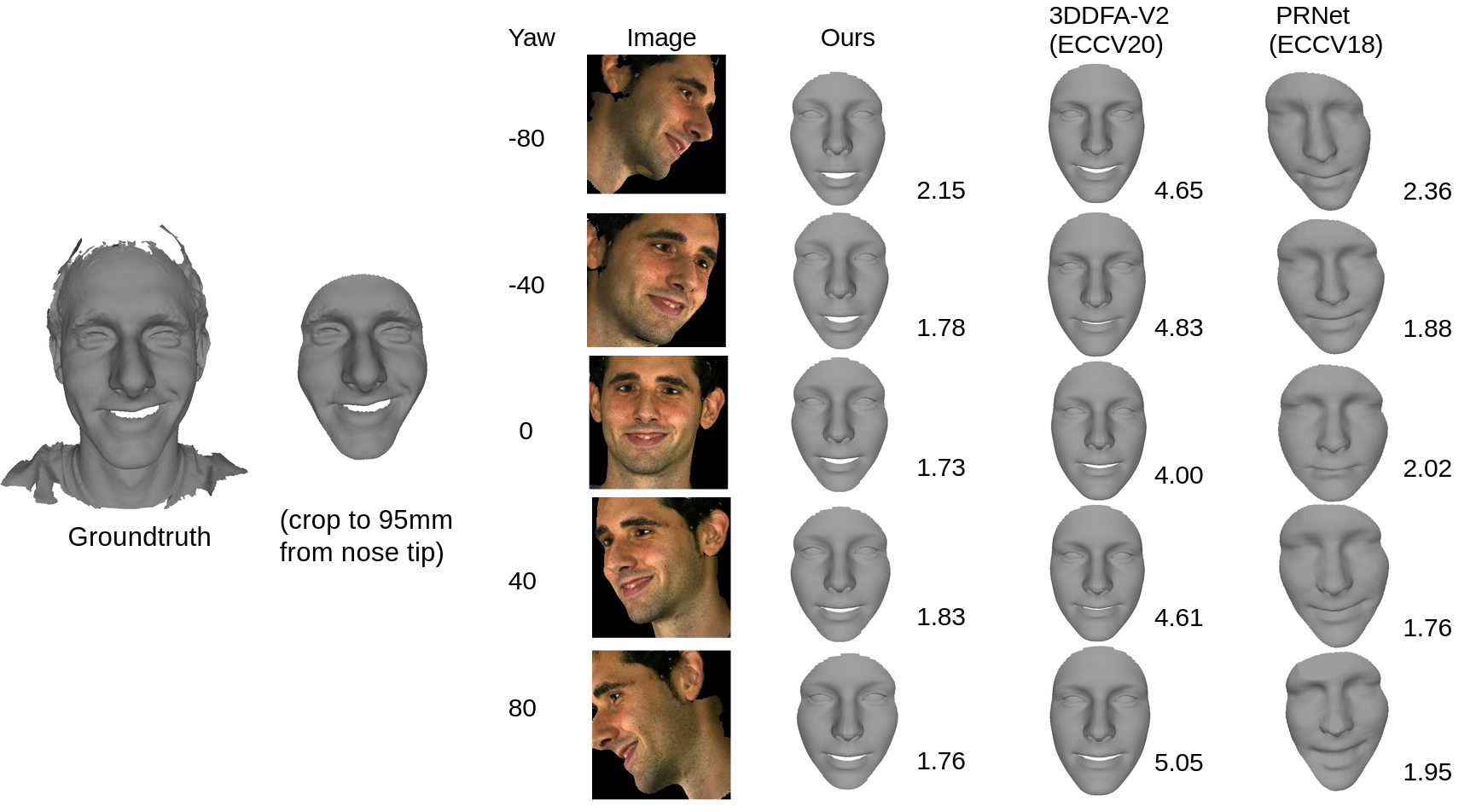}
    \includegraphics[width=0.85\linewidth]{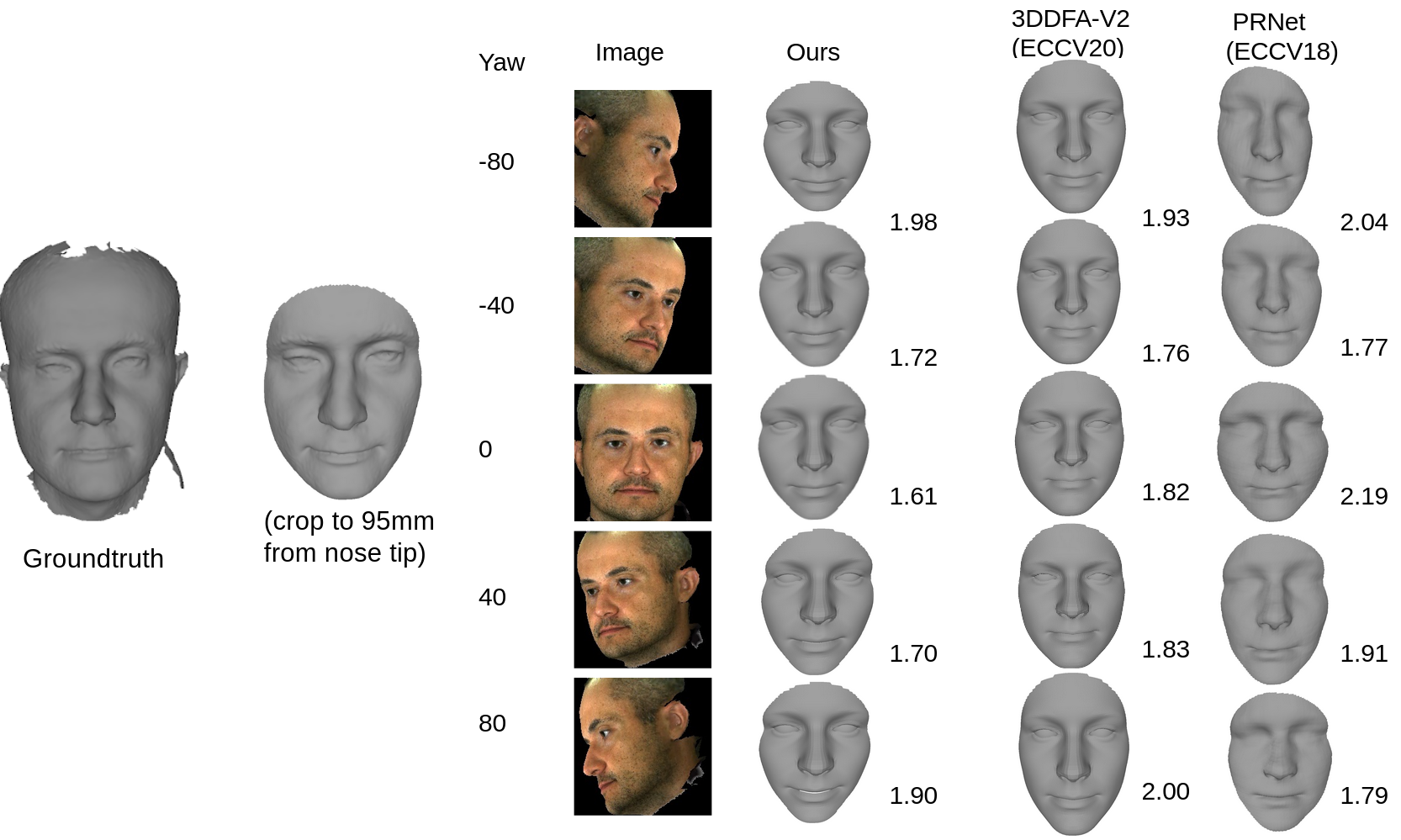}
    \caption{\textbf{Reconstructed face comparison by yaw angle on examples from the Florence dataset.} Numbers beside the reconstructed models are their normalized point-to-plane RMSEs. Our results are robust to pose changes. For the upper example, 3DDFA-V2 shows wider faces, unapparent cheeks, non-pointed chins, and larger forehead areas with a cropping range of 95mm from the nose tip; therefore, their results hold higher errors. PRNet shows imprecise facial structures. In addition, their faces are twisted for large pose cases. For the lower example, the groundtruth face is wider, but 3DDFA-V2 shows more elongated faces, and PRNet predictions are unreliable. Our results are more similar to the groundturh shape.}
    \label{florence_recon_comp}
\end{figure*}

The introduced UV-texture GAN adopts a pixel-to-pixel image translator that transforms unstructured in-the-wild images to a canonical UV space to generate the UV-texture maps from images. However, pixel-to-pixel style transfer \cite{isola2017image, wang2018pix2pixHD} retains input image structures, such as salient object outlines, and produces a different style artifact. It is hard to map unconstrained face images onto the canonical UV space by direct pixel-to-pixel translators. To resolve the issue, we further feed a template UV map together with a face image as inputs to the generator (Fig. \ref{uvgan}). The template is projected from the mean texture of BFM Face. Further, we shortcut the template to the output for facilitating the training procedure, where the generator learns a mapping from the six-channel input to the residual UV space. We display the ability of the template in Fig. \ref{template_compare}.

To form our training set, we collect about 2K in-the-wild frontal face images and warp the faces onto the UV space with the aids of facial landmarks. The generator and discriminator architectures are the same as pix2pix model \cite{isola2017image}. Least-square GAN (LSGAN) \cite{mao2017least} is used as the loss for training. We train the network with 300 epochs. Adam is adopted as the optimizer with an initial learning rate of 0.0002. After 100 epochs, the learning rate starts to drop linearly to 0.

We show a comparison in Fig.\ref{uv_compare} for synthesized textures from the introduced UV-texture GAN and conventional 3DMM texture fitting. Results of UV-texture GAN are more realistic and are not over-smooth compared with textures from 3DMM fitting. Skin colors are more similar to images since the introduced UV-texture GAN combines hues from images and structures from the template to produce more realistic textures. 

\section{More Qualitative Results}
\label{i}
Here we further show more qualitative results on the 300VW dataset for talks or interview videos \cite{shen2015first} in Fig. \ref{300vw-1}, \ref{300vw-2}, \ref{300vw-3}, \ref{300vw-4} and Artistic Faces (AF) for different artistic style faces \cite{yaniv2019face} in Fig. \ref{af-1}, \ref{af-2}.

\begin{figure*}[bt!]
    \centering
    \includegraphics[width=1.0\linewidth]{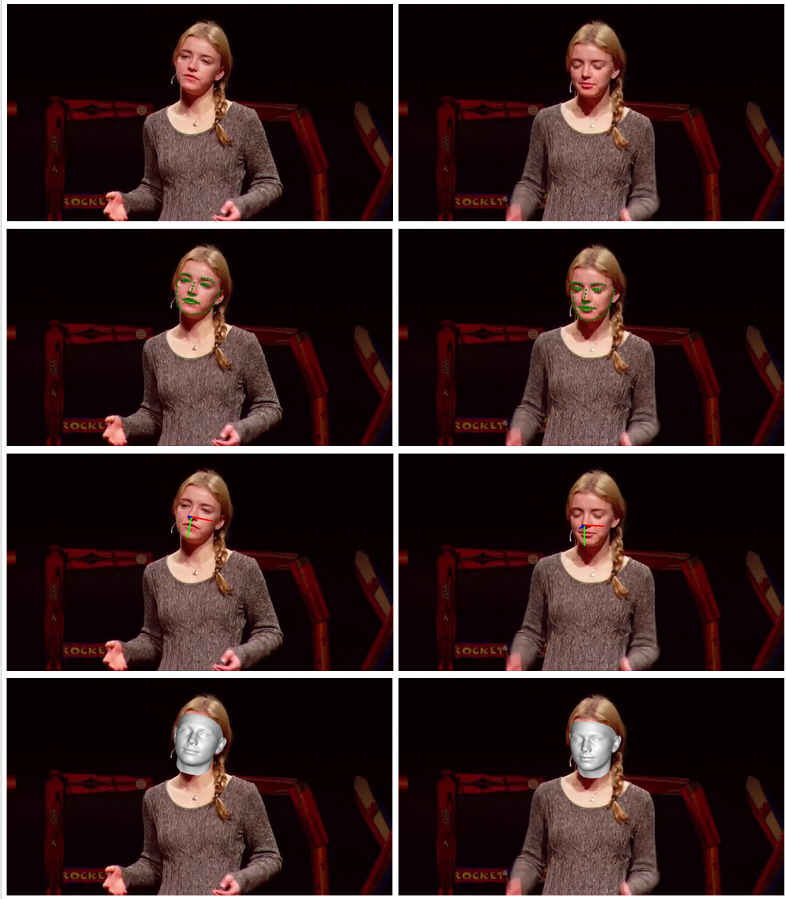}
    \caption{\textbf{Results of 3D geometry prediction on 300VW from our method.} Row 1-4: images, 3D landmarks, face orientation, 3D faces.}
    \label{300vw-1}
\end{figure*}

\begin{figure*}[bt!]
    \centering
    \includegraphics[width=1.0\linewidth]{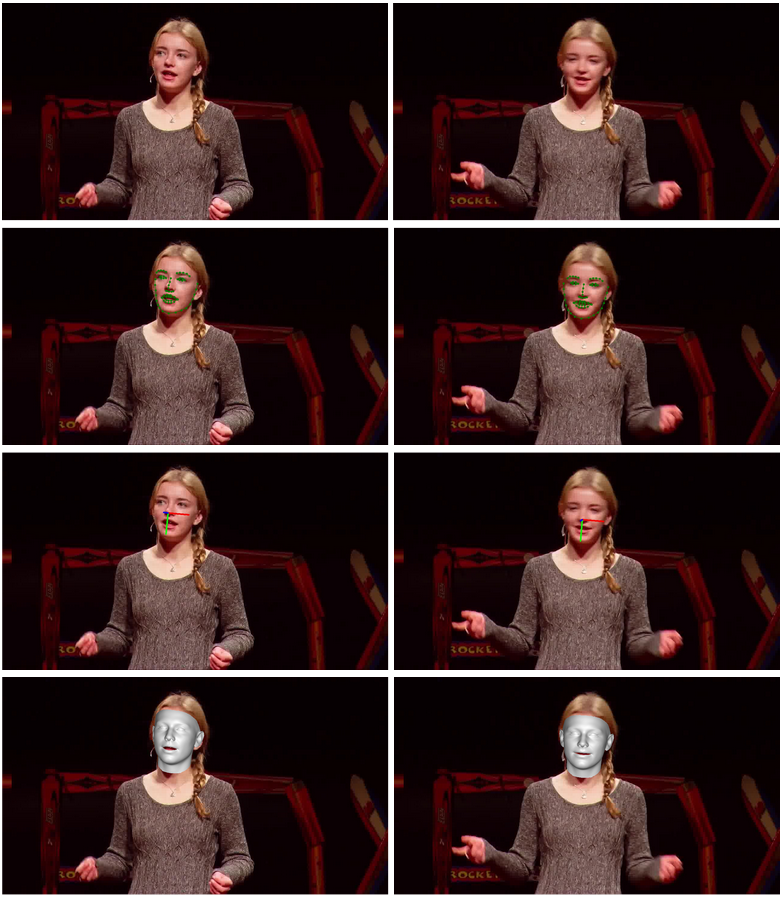}
    \caption{\textbf{(Continued) Results of 3D geometry prediction on 300VW from our method.} Our result is robust to motion blur for the right-hand-side case.}
    \label{300vw-2}
\end{figure*}

\begin{figure*}[bt!]
    \centering
    \includegraphics[width=1.0\linewidth]{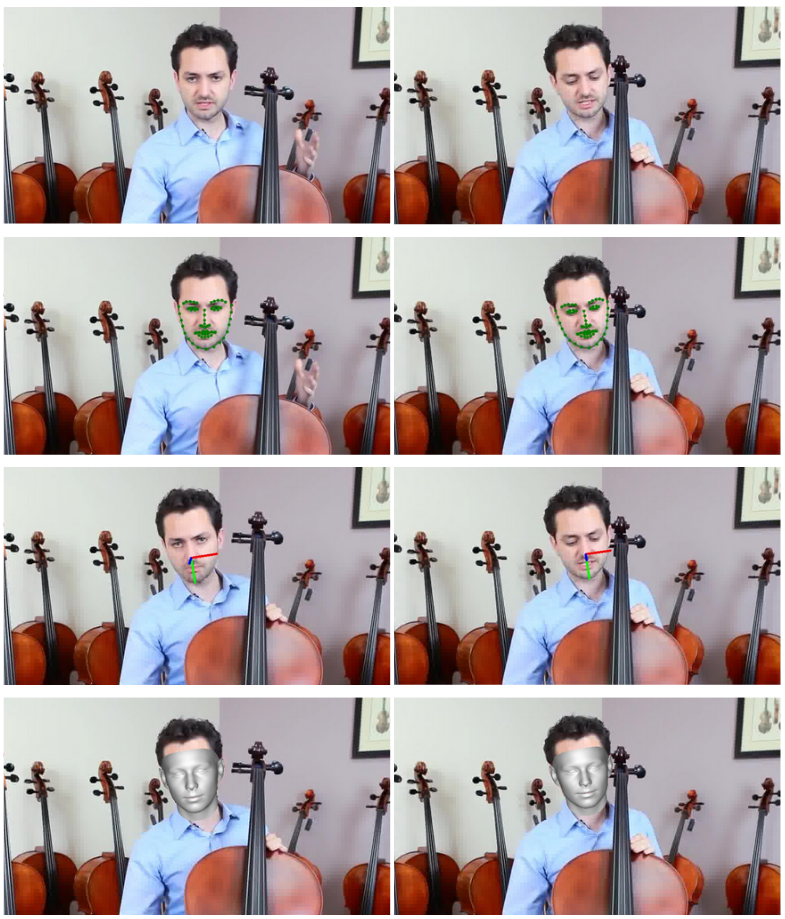}
    \caption{\textbf{(Continued) Results of 3D geometry prediction on 300VW from our method.}}
    \label{300vw-3}
\end{figure*}

\begin{figure*}[bt!]
    \centering
    \includegraphics[width=1.0\linewidth]{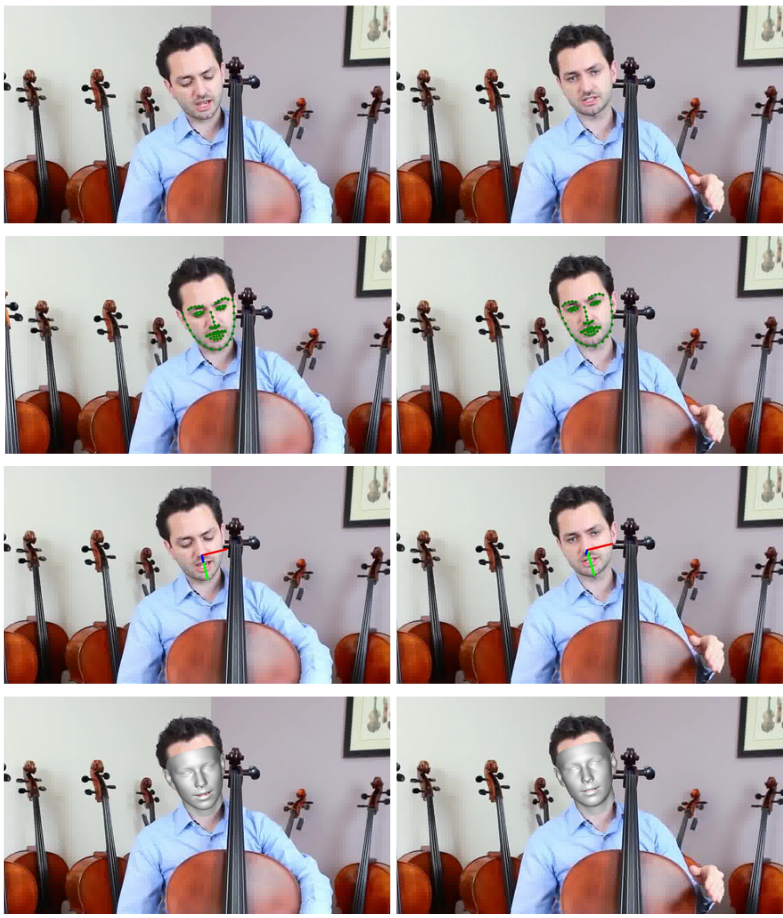}
    \caption{\textbf{(Continued) Results of 3D geometry prediction on 300VW from our method.}}
    \label{300vw-4}
\end{figure*}

\begin{figure*}[bt!]
    \centering
    \includegraphics[width=1.0\linewidth]{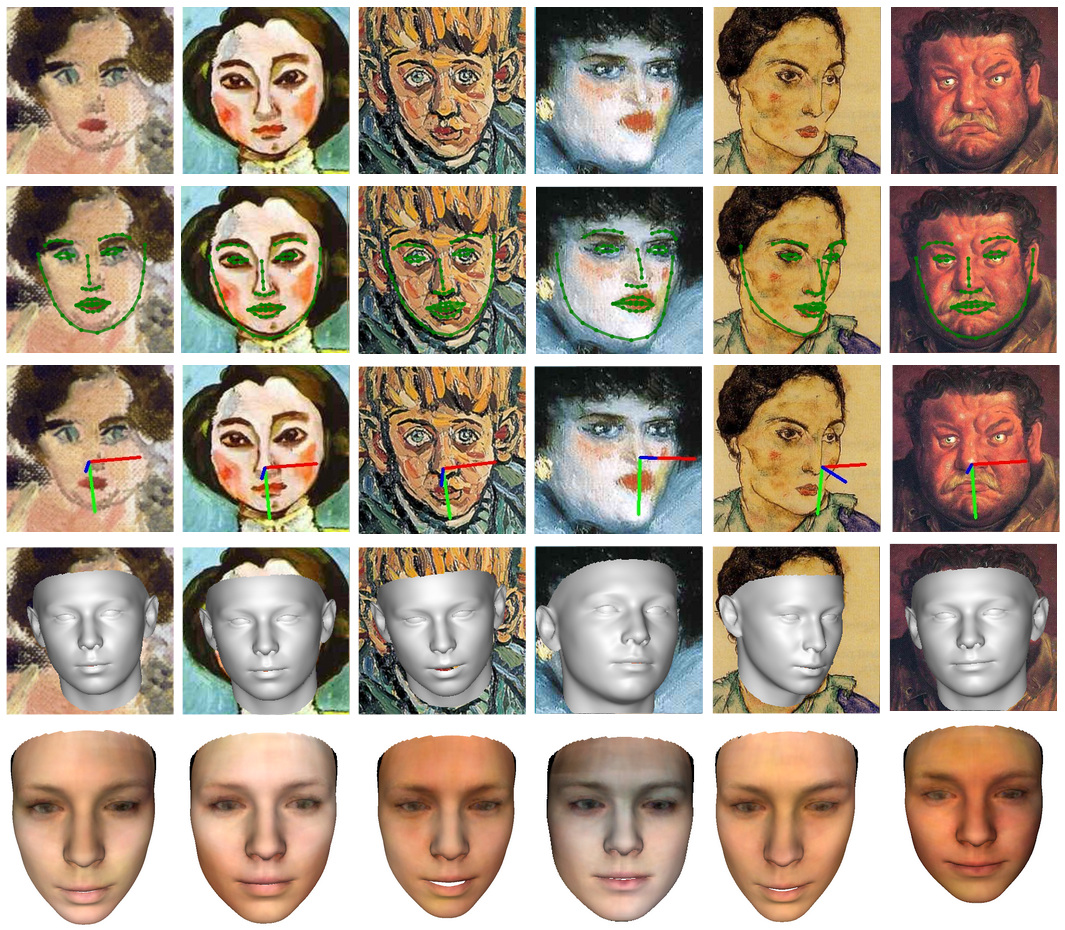}
    \caption{\textbf{Results of 3D geometry prediction on Artistic Faces from our method.} Row 1-5: images, 3D landmarks, face orientation, 3D faces, textures.}
    \label{af-1}
\end{figure*}

\begin{figure*}[bt!]
    \centering
    \includegraphics[width=1.0\linewidth]{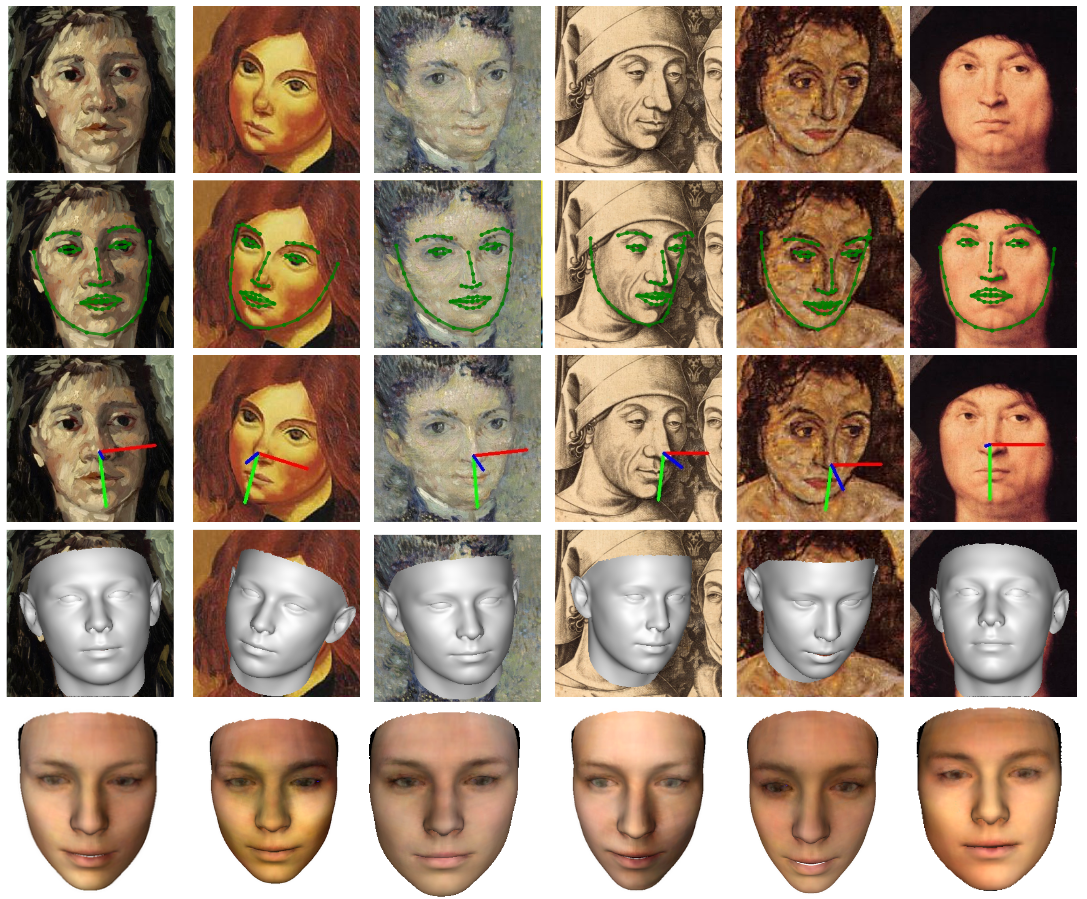}
    \caption{\textbf{(Continued) Results of 3D geometry prediction on Artistic Faces from our method.}}
    \label{af-2}
\end{figure*}

\end{document}